\documentclass[lettersize,journal]{IEEEtran}
\usepackage{amsmath,amsfonts}
\usepackage{algorithmic}
\usepackage{algorithm}
\usepackage{array}
\usepackage[caption=false,font=normalsize,labelfont=sf,textfont=sf]{subfig}
\usepackage{textcomp}
\usepackage{stfloats}
\usepackage{url}
\usepackage{verbatim}
\usepackage{graphicx}
\usepackage{cite}
\usepackage{xcolor}
\usepackage{multirow}
\usepackage{wrapfig}
\usepackage{paralist}
\usepackage{hyperref}
\usepackage{orcidlink}

\begin{document}

\title{RBFleX-NAS: Training-Free Neural Architecture \\ Search Using Radial Basis Function Kernel and \\ Hyperparameter Detection}
\author{
    Tomomasa Yamasaki,~\IEEEmembership{Student Member,~IEEE}
    \textsuperscript{\orcidlink{0000-0002-5610-8534}},
    Zhehui Wang \textsuperscript{\orcidlink{0000-0002-7139-724X}},
    Tao Luo,~\IEEEmembership{Member,~IEEE} \textsuperscript{\orcidlink{0000-0002-3415-3676}}, \\
    Niangjun Chen,
    and Bo Wang,~\IEEEmembership{Senior Member,~IEEE} \textsuperscript{\orcidlink{0000-0001-9199-0799}
    }
}




\maketitle
\begin{abstract}
\label{sec:abstract}
Neural Architecture Search (NAS) is an automated technique to design optimal neural network architectures for a specific workload. Conventionally, evaluating candidate networks in NAS involves extensive training, which requires significant time and computational resources. To address this, training-free NAS has been proposed to expedite network evaluation with minimal search time. However, state-of-the-art training-free NAS algorithms struggle to precisely distinguish well-performing networks from poorly-performing networks, resulting in inaccurate performance predictions and consequently sub-optimal top-1 network accuracy. Moreover, they are less effective in activation function exploration.
To tackle the challenges, this paper proposes RBFleX-NAS, a novel training-free NAS framework that accounts for both activation outputs and input features of the last layer with a Radial Basis Function (RBF) kernel. 
We also present a detection algorithm to identify optimal hyperparameters using the obtained activation outputs and input feature maps. We verify the efficacy of RBFleX-NAS over a variety of NAS benchmarks. RBFleX-NAS significantly outperforms state-of-the-art training-free NAS methods in terms of top-1 accuracy, achieving this with short search time in NAS-Bench-201 and NAS-Bench-SSS. In addition, it demonstrates higher Kendall correlation compared to layer-based training-free NAS algorithms.
Furthermore, we propose NAFBee, a new activation design space that extends the activation type to encompass various commonly used functions. In this extended design space, RBFleX-NAS demonstrates its superiority by accurately identifying the best-performing network during activation function search, providing a significant advantage over other NAS algorithms.
\end{abstract}

\begin{IEEEkeywords}
Training-free, neural architecture search, Radial Basis Function, activation
\end{IEEEkeywords}

\section{Introduction}
\label{sec:introduction}
\IEEEPARstart{O}{ver} the past decade, Deep Neural Networks (DNNs) are intensively utilized in numerous Artificial Intelligence (AI) applications across a multitude of domains
in computer vision\cite{10085276}, natural language processing\cite{devlin2019bert}, social network analysis\cite{fan2019graph}, recommendation systems\cite{9408167}, etc. However, deploying DNNs to achieve high accuracy for a specific task can incur substantial time and effort when manually designing the network and identifying the optimal architectures and hyperparameters. Neural Architecture Search (NAS) has been proposed to address this challenge.
Conventional NAS automates the process of architecture search by generating candidate networks and validating their accuracy with training \cite{NAS-Train1, NAS-Train2, NAS-Train3, 9496596, 9060902, 9956879, 9646995, 9211549}. However, this approach is computationally expensive as network training can take a vast amount of time, particularly when network architectures become deep and complex. 
For instance, a NAS algorithm leveraging reinforcement learning can take 28 days with 800 GPUs to identify a well-performed DNN topology \cite{NAS-Train1}.
To speed up NAS and reduce computational cost, researchers have proposed various techniques. 
LightNAS \cite{9896156} estimates candidate network potential based on the total training losses for several epochs, which can reduce the training cost.
Another efficient approach is leveraging neural predictors with simple regression models rather than training actual networks.
The models can adopt random forest\cite{E2EEP}, auto-encoder\cite{SSANA}, or decision tree\cite{9698855} method to forecast the network performance. However, the regression models in the neural predictors still require training to ensure good performance. 

To mitigate the issue, gradient-based zero-cost proxies for NAS have been proposed to evaluate networks in their initial state without the need for full training. This method evaluates networks by analyzing the knowledge obtained from forward or backward propagation. One basic proxy is grad\_norm \cite{abdelfattah2021zerocost}, which aggregates the Euclidean norm of the gradients from a single minibatch of training data. Snip proxy\cite{lee2018snip} evaluates networks using a saliency metric computed at initialization based on a single minibatch of data, approximating the change in loss (e.g., cross-entropy loss) when a specific parameter is removed. Unlike snip, synflow proxy\cite{NEURIPS2020_46a4378f} simply calculates its loss as the product of all network parameters without using minibatch data and cross-entropy loss. \textcolor{black}{ZiCo\cite{li2023zico} estimates training convergence
speed and generalization capacity based on the mean value and standard deviation of the gradients. EProxy\cite{li2023extensible} leverages a few-shot self-supervised
regression to evaluate networks.} Nonetheless, these gradient-based proxies are only effective for a restricted set of design spaces.
In contrast, layer-based approaches do not require labels or loss values to predict the trained accuracy of a network, thus completely eliminating the training cost. For example, 
\textcolor{black}{TE-NAS\cite{chen2021neural} analyzes the spectrum of the neural tangent kernel and the number of linear regions in the input space to score a network.} NASWOT\cite{NASWOT} provides a scoring system based on the difference in ReLU output among minibatches to predict the final trained accuracy, while DAS\cite{DAS} further decouples the NASWOT scoring framework into distinguishing atomic and activation atomic metrics.

\IEEEpubidadjcol

\begin{figure}
\centering
\includegraphics[width=0.45\textwidth]{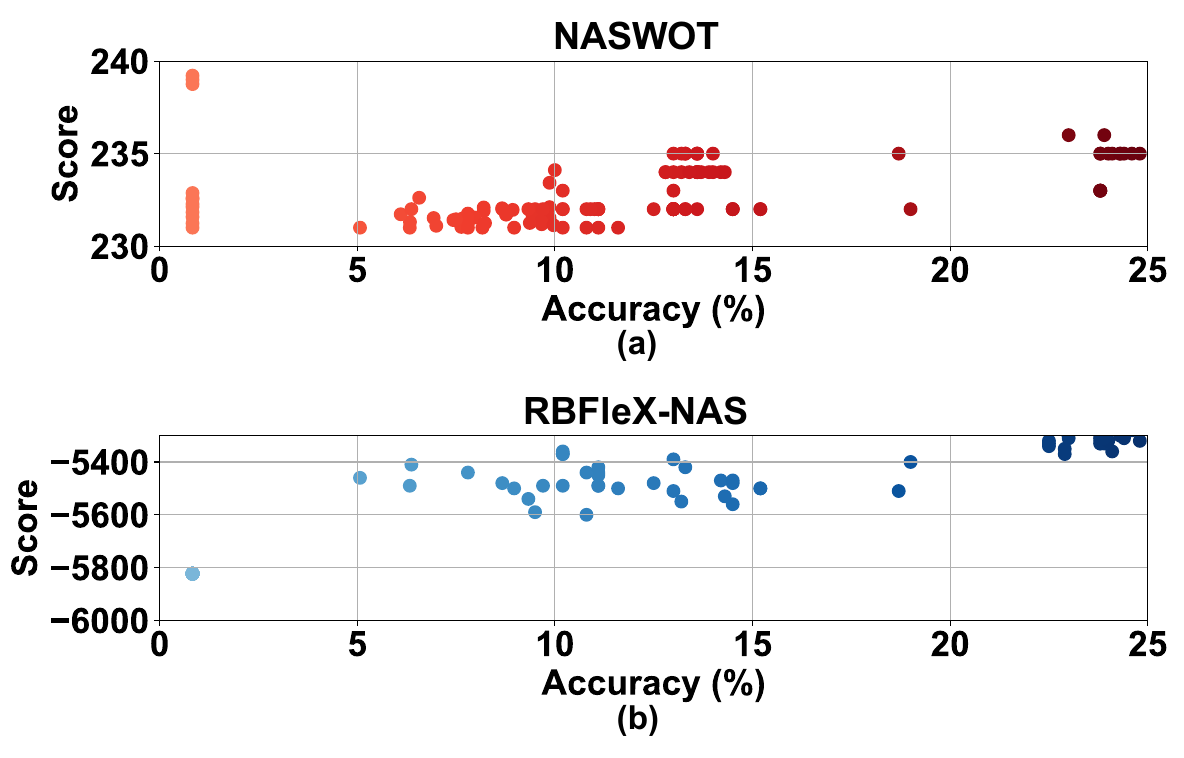}
\caption{The score vs. accuracy from (a) NASWOT and (b) proposed RBFleX-NAS when testing with NAS-Bench-201 over ImageNet. NASWOT cannot distinguish the network performance on the basis of their scores when the ground-truth accuracy on the ImageNet dataset falls below 25\%. Note that the lighter the color, the lower the accuracy.} 
\label{fig:error_NASWOT}
\end{figure}

We propose RBFleX-NAS, a novel training-free NAS algorithm that further improves the scoring performance and extends the network design space to various activation functions.
Our algorithm scores an individual network by examining the similarity of the activation outputs and the similarity of the input feature maps of the last layer across different input images. This is because the last layer can play a key role in projecting the extracted features into lower dimensions by acting as a classifier \cite{yosinski2014transferable}.
Specifically, we exploit the Radial Basis Function (RBF) kernel to derive the similarity among images in a minibatch and predict the final trained accuracy of the network. Simultaneously, we develop a Hyperparameter Detection Algorithm (HDA), which searches for appropriate hyperparameters to assist the RBF kernels. \textcolor{black}{We have validated that RBFleX-NAS achieves significantly higher Kendall correlation between scores and final trained accuracies of networks across widely used NAS benchmarks, including NAS-Bench-201\cite{dong2020nasbench201}, NATS-Bench-SSS\cite{NATS-Bench}, NDS\cite{NDS}
as well as TransNAS-Bench-101\cite{duan2021transnas}.}
Our source code is available at \textit{https://github.com/Edge-AI-Acceleration-Lab/RBFleX-NAS.git}

Our contributions are summarized as follows.
\begin{itemize}
\vspace{11pt}
\item We propose RBFleX-NAS, a zero-cost NAS algorithm that leverages the similarity of activation outputs and the similarity of the input feature maps of the last layer among input images. RBFleX-NAS is more precise in distinguishing network performance with scores, achieving higher top-1 accuracy in search. 
\vspace{11pt}
\item We propose NAFBee, a new design space for benchmarking purposes with VGG-19 and BERT, further extending the activation type from ReLU to various commonly used functions. RBFlex-NAS successfully identifies the best-performing network with activation function search, which the current NAS algorithms fail to do.
\vspace{11pt}
\item We achieve significantly superior accuracy correlation in network scoring compared to state-of-the-art layer-based training-free NAS across various NAS benchmarks. 
\end{itemize}

\section{Prior Arts}
\label{sec:related}
A variety of techniques have been developed for network architecture search, such as reinforcement learning\cite{NAS-Train1, NAS-Train3}, evolutionary algorithm\cite{NAS_EA2}, gradient algorithm\cite{NAS_GA1}, Deep Q Network (DQN)\cite{NAS_DQN1}, etc. However, they all have to train the candidate networks for performance evaluation, rendering a large amount of computation time and GPU power.

Recently, layer-based NAS algorithms without training have been proposed to address the problem. The algorithms are completely training-free and thus do not need labels or loss information. For instance, NASWOT\cite{NASWOT} and DAS\cite{DAS} score candidate networks and correlate the scores to their final trained accuracy rather than train them directly. NASWOT evaluates the networks based on the difference in ReLU output among the images in a minibatch. 
Specifically, ReLU partitions the input feature maps into either an activated region when the input is positive or an inactivated region otherwise. NASWOT treats these regions as binary features and scores a network accordingly. When a network exhibits a substantial difference in ReLU output among the input images, NASWOT recognizes it as a well-performed architecture as it indicates that the network can respond well to the features of different images. DAS further decouples the score equation in NASWOT into a distinguishing atomic and an activation atomic. The distinguishing atomic capture the network's capability to differentiate among different inputs, while the activation atomic takes the number of activation functions to simplify the comparison among networks. Compared to NASWOT, DAS runs faster with higher reliability for certain network design spaces.

\begin{figure}
\centering
\includegraphics[width=0.45\textwidth]{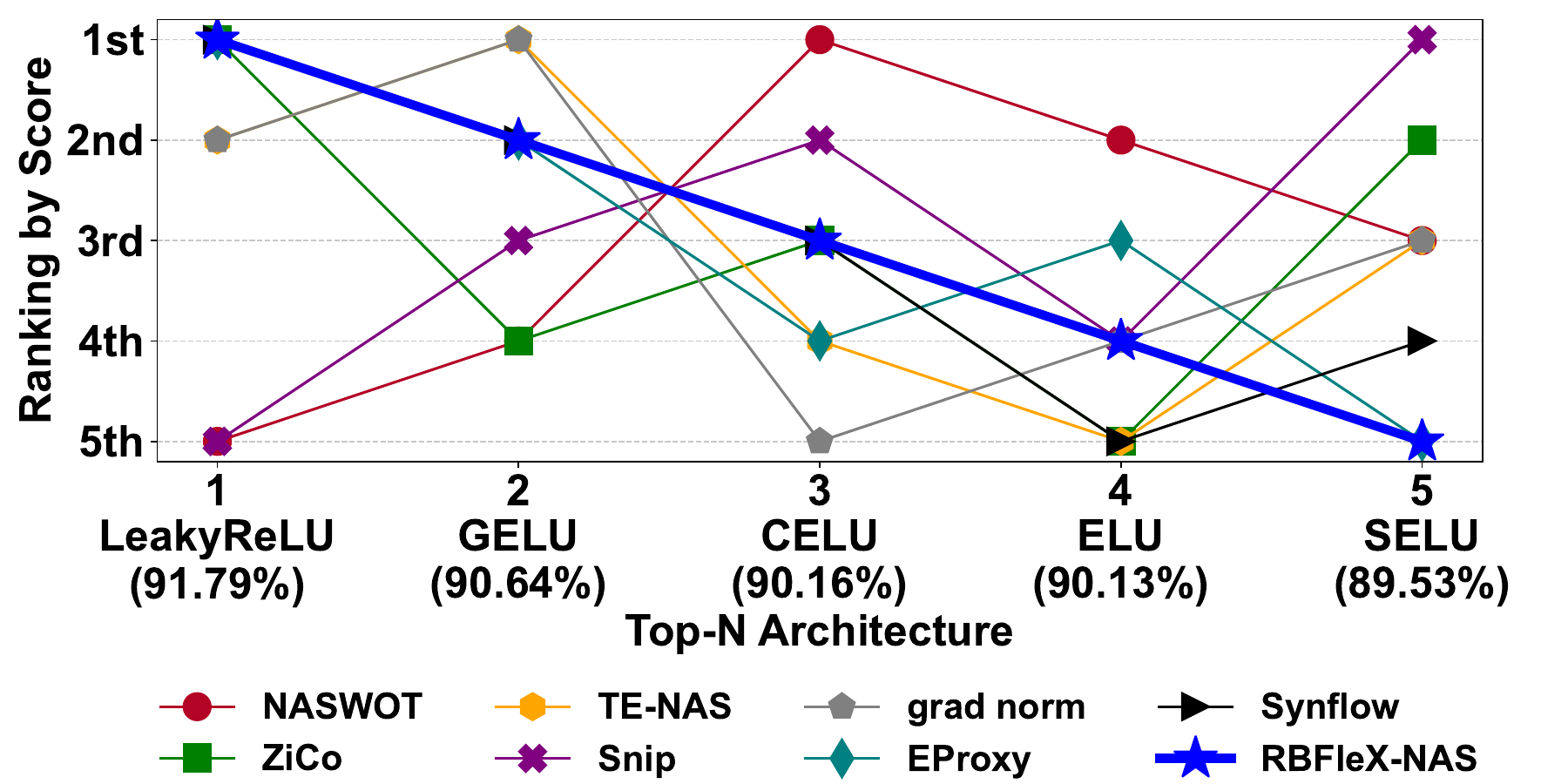}
\caption{\textcolor{black}{Illustration of activation function exploration on VGG-11 with CIFAR-10 dataset. The Y-axis represents the accuracy ranking based on the score, which is evaluated by various training-free NAS methods using networks designed with five distinct non-ReLU activation functions.}
}
\label{fig:ADE}
\end{figure}


However, NASWOT and DAS impose two limitations in scoring and design space. In the scoring phase, NASWOT and DAS merely check the ReLU output and the number of ReLU layers. However, the ReLU-related information is insufficient to predict the final accuracy of a network \cite{yosinski2014transferable}. For instance, NASWOT can not distinguish the network performance based on their scores
when the ground-truth accuracy of the ImageNet dataset falls below 25\% (Fig. \ref{fig:error_NASWOT}(a)). In particular, the two networks with an accuracy of 0.8\% obtain even higher scores compared to the remaining candidates, manifesting inaccurate performance predictions by NASWOT. 
This is because the last layer (e.g., a Fully Connected layer with softmax) also plays a key role in projecting the extracted features into lower dimensions by acting as a classifier \cite{yosinski2014transferable}. 
\textcolor{black}{Moreover, Fig. \ref{fig:ADE} reveals that NASWOT exhibits a poor correlation between score and network accuracy across different activation functions. Similarly, other state-of-the-art training-free NAS algorithms, including TE-NAS, grad\_norm, Synflow, Snip, ZiCo, and EProxy perform poorly in differentiating network accuracy attributed to different activation functions.} 

Hereby, we propose RBFleX-NAS, a novel algorithm that leverages RBF kernels with hyperparameter detection to forecast the final trained accuracy of the candidate networks by taking the last layer into account. 
Moreover, it enables activation function exploration in the design space, offering more flexibility for DNN design.

\section{Method}
\label{sec:method}
This section presents a detailed description of RBFleX-NAS for the network search. We begin by discussing our kernel-based approach and explaining how to measure the similarity. Next, we introduce HDA to obtain the fine-tuned hyperparameters for the RBF kernel. 
Subsequently, we elaborate on network initialization and the effect of minibatch size on determining a score. Finally, we discuss the impact of images in different batches. 
\subsection{Analyzing Networks using RBF Kernel}
\label{sec:RBFNAS}

\begin{figure}
\centering
\includegraphics[width=0.48\textwidth]{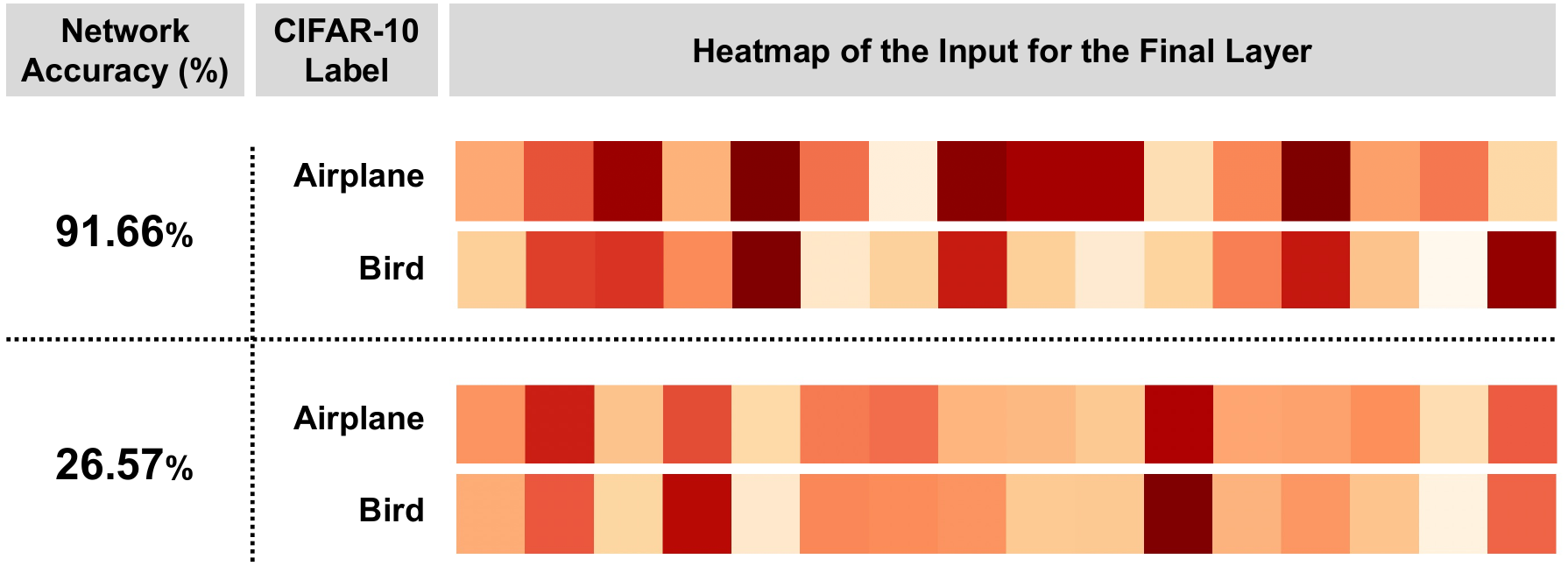} 
\caption{\textcolor{black}{Input heatmap of the last layer from a network with higher accuracy (91\%) and a network with lower accuracy (26\%) over CIFAR-10. }}
\label{fig:Discussion}
\end{figure}

\textcolor{black}{A critical factor contributing to the performance of RBFleX-NAS is its ability to analyze the full information of the feature maps in the network layers by RBF kernels with proposed HDA, as opposed to relying on the binary representations utilized by state-of-the-art layer-based methods such as NASWOT and TE-NAS. RBFleX-NAS enables a more detailed evaluation of the activation function outputs while simultaneously considering the inputs to the final layer, which are critical for determining the network’s performance. The heatmap in Fig. \ref{fig:Discussion} illustrates that networks with higher accuracy can generate more distinct input representations for the final layer compared to networks with lower accuracy. RBFleX-NAS effectively captures the critical distinctions, enabling it to more accurately evaluate and rank network performance.}

NASWOT \cite{NASWOT} reveals that a well-performed network architecture tends to have a lower similarity in terms of ReLU outputs among different input images.
However, it is worth noting that a network can still exhibit low accuracy though its ReLU output shows low similarity. This is because the network performance is indicated not only by the activation layers but also by the last layer \cite{yosinski2014transferable}.

\begin{figure}
\centering
\includegraphics[width=0.35\textwidth]{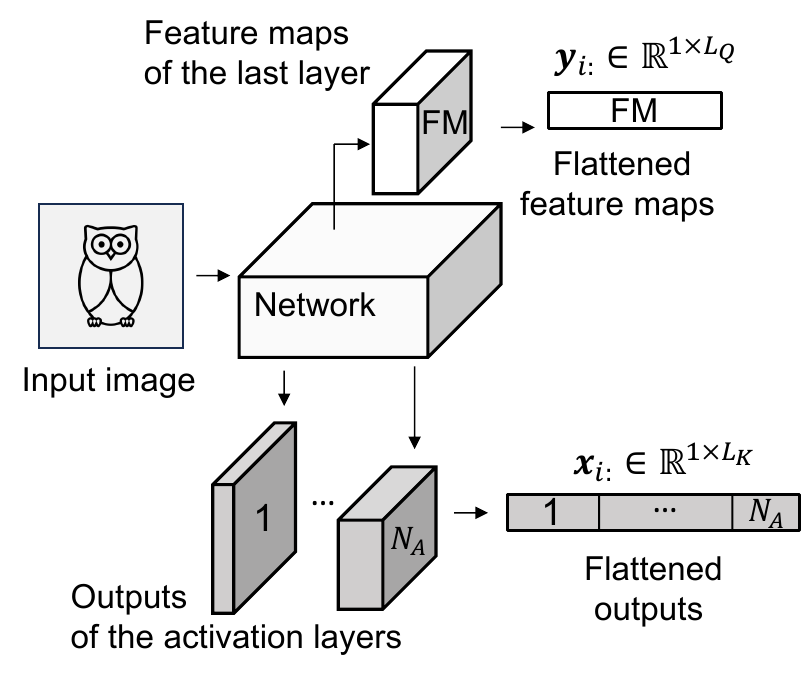}
\caption{Vector processing on RBFleX-NAS, where $N_A$ is the number of activation functions.}
\label{fig:ch2_Getarray}
\end{figure}

In this section, we score a network architecture by accounting for both the similarity among the activation outputs and the similarity among the inputs of the last layer. Fig. \ref{fig:ch2_Getarray} depicts the process of obtaining the output vector of the activation layer, ${\bf x}_{i:}\in \mathbb{R}^{1\times L_K}$ and the input feature map vector of the last layer, ${\bf y}_{i:}\in \mathbb{R}^{1\times L_Q}$, where $L_K$ and $L_Q$ denote the length of these vectors, respectively. We flatten the activation output and input feature map vector for each input image. Specifically, we convert each three-dimensional output (i.e., width$\times$height$\times$channels) from an activation function into a one-dimensional vector and concatenate them horizontally to obtain the output vector. Similarly, we generate the input feature map vector by transforming the input feature maps into one dimension.

\begin{figure*}
\centering
\includegraphics[width=0.8\textwidth]{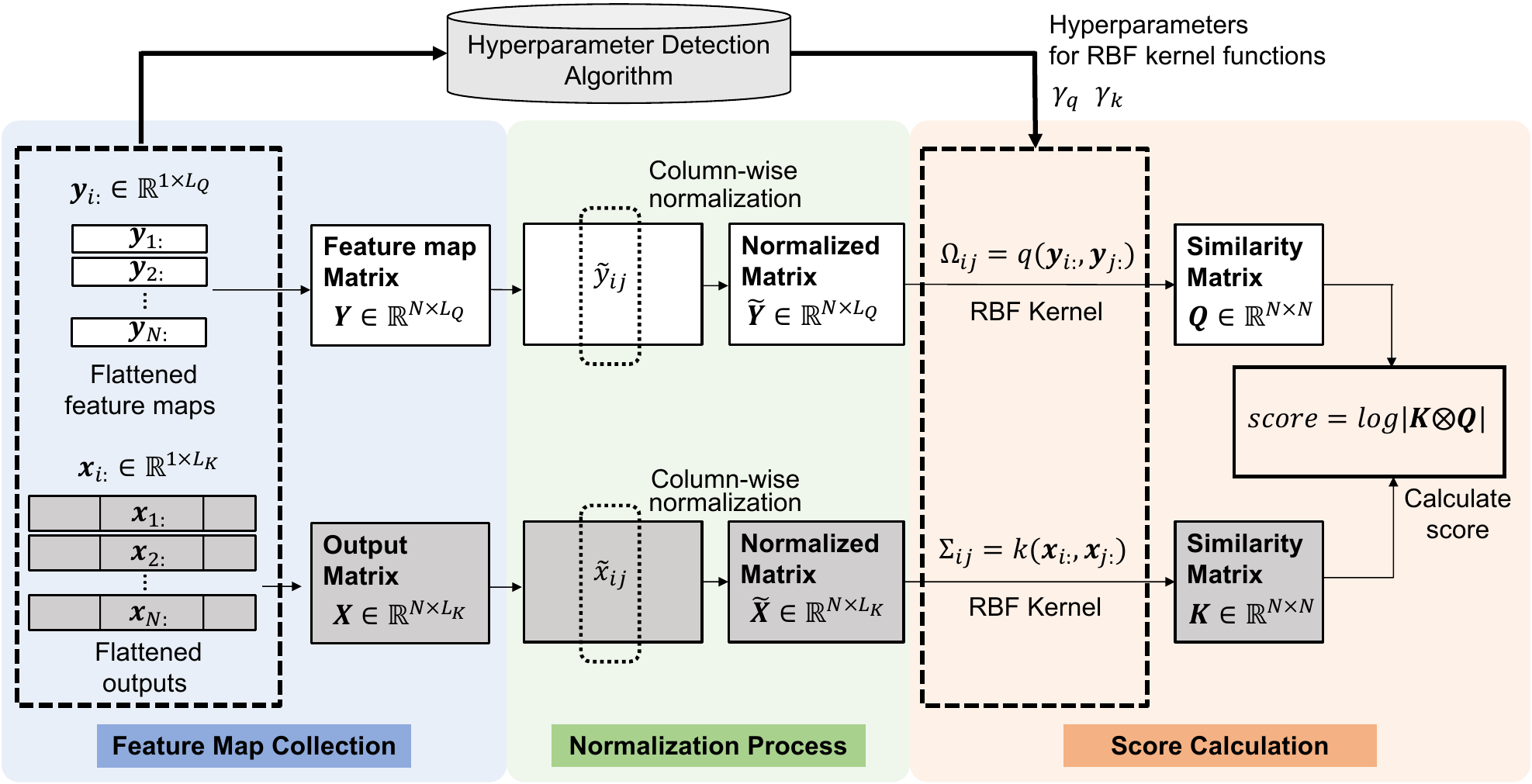}
\caption{The score computation process in RBFleX-NAS. 
}
\label{fig:ch2_score}
\end{figure*}

Fig. \ref{fig:ch2_score} shows the scoring process in our proposed RBFleX-NAS. Firstly, the framework generates $N$ output vectors and $N$ input feature map vectors based on the $N$ input images in a minibatch. RBFleX-NAS then concatenates them vertically as an output matrix, ${\bf X}\in \mathbb{R}^{N\times L_K}$ and an input feature map matrix, ${\bf Y}\in \mathbb{R}^{N\times L_Q}$, respectively. To better measure the similarity among different images, we normalize the two matrices with a range of 0 $\sim$ 1. There are three common matrix normalization techniques, namely element-wise normalization,
row-wise normalization, and column-wise normalization, respectively. However, we observe the element-wise normalization and the row-wise normalization reduce the data correlation among rows in ${\bf X}$ or ${\bf Y}$. Therefore, we adopt the column-wise normalization and illustrate the method with Equation (\ref{norm}).
\begin{equation}
    \label{norm}
    \tilde{\alpha}_{ij}=
    \begin{cases}
        \frac{\alpha_{ij}-min(\alpha_{:j})}{max(\alpha_{:j})-min(\alpha_{:j})} & \text{if $max(\alpha_{:j}) \ne min(\alpha_{:j})$} \\
        \alpha_{ij} & \text{otherwise}
    \end{cases}
\end{equation}

\noindent where $\tilde{\alpha}_{ij}$ is the value after normalization, $\alpha_{ij}$ is an element in $\bf X$ or $\bf Y$ and $\alpha_{:j}$ denotes the $j$-th column vector of $\bf X$ or $\bf Y$.

After obtaining the normalized matrices $\tilde{\bf X}$ and $\tilde{\bf Y}$, we calculate the similarity matrices of the activation output $\bf K\in \mathbb{R}^{N\times N}$ and the input feature maps $\bf Q\in \mathbb{R}^{N\times N}$ by leveraging RBF kernel. 

The RBF kernel is a non-linear function whose elements are representations of distance from the center, which is widely utilized in Support Vector Machine \cite{svm1}\cite{svm2}, RBF networks \cite{rbfnet1}, multi-task Gaussian process \cite{mgp1}, etc as it can perform better in extracting complex features than many other kernels \cite{vert2004primer}. 
However, the RBF kernel is highly sensitive to the choice of hyperparameters. To address this issue, we develop an HDA which will be elaborated in Section \ref{sec:self-detectH}. Thanks to the HDA, we can obtain fine-tuned hyperparameters (i.e., $\gamma_k$ and $\gamma_q$) to generate a precise similarity matrix. Specifically, each term in the similarity matrices (i.e., $\Sigma_{ij}$ and $\Omega_{ij}$) is calculated with the RBF kernel as Equations (\ref{RBF_K}) and (3) describe.
\begin{align}
    \label{RBF_K}
    \Sigma_{ij} =\exp({-\gamma_k\|\tilde{\bf x}_{i:}-\tilde{\bf x}_{j:}\| ^2}) \\
    \Omega_{ij} = \exp({-\gamma_q\|\tilde{\bf y}_{i:}-\tilde{\bf y}_{j:}\| ^2})
\end{align}

Therefore, the similarity matrix of the activation outputs, denoted by $\bf K\in \mathbb{R}^{N\times N}$, and the similarity matrix of the last input feature maps, denoted by $\bf Q\in \mathbb{R}^{N\times N}$, can be represented as follows.
\begin{align}
    \label{SM_K}
    \bf K = 
    \begin{pmatrix}
     \Sigma_{11} & \cdots &\Sigma_{1N}  \\
     \vdots & \ddots & \vdots  \\
     \Sigma_{N1}  & \cdots &\Sigma_{NN}  
    \end{pmatrix}
\end{align}

\begin{align}
    \label{SM_Q}
    \bf Q = 
    \begin{pmatrix}
     \Omega_{11} & \cdots & \Omega_{1N}  \\
     \vdots & \ddots & \vdots  \\
     \Omega_{N1} & \cdots & \Omega_{NN}
    \end{pmatrix}
\end{align}

\noindent A small value of $\Sigma_{ij}$ or $\Omega_{ij}$ reveals that the similarity is low between the $i$-th vector and the $j$-th vector in $\tilde{\bf X}$ and $\tilde{\bf Y}$, respectively. Furthermore, a potentially good network tends to possess low similarity in both ${\bf K}$ and ${\bf Q}$ among various input images.
As the matrices ${\bf K}$ and ${\bf Q}$ denote different similarities from the perspective of distinct layers, we exploit the Kronecker product $\otimes$ to integrate the two matrices into one Figure-of-Merit for our scoring process. Specifically, we score a given network with the Kronecker product as Equation (\ref{score}) shows. A higher score indicates a network with a higher accuracy.
\begin{align}
    \label{score}
    score = \log{|{\bf K\otimes \bf Q}|}
\end{align}

\noindent Here we can further simplify Equation (\ref{score}) to
\begin{align}
    \label{score_v2}
    score = N*(\log|{\bf K}| + \log|{\bf Q}|)
\end{align}

\noindent Consequently, the computational complexity of the score calculation becomes $O(N^{2.373})$ according to the method in \cite{aho1974design}.

\begin{figure}
\includegraphics[width=0.48\textwidth]{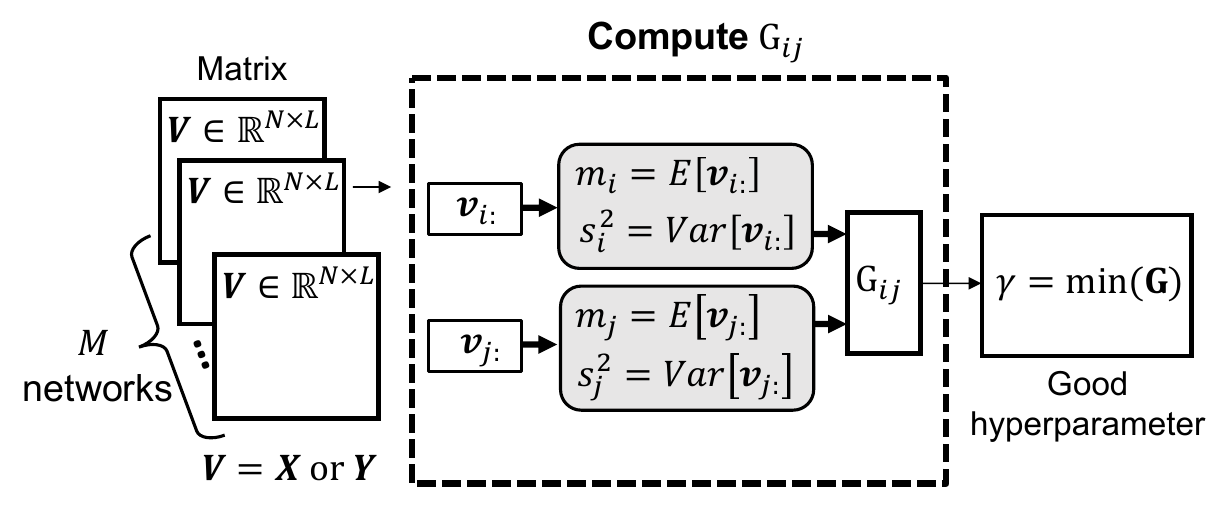}
\caption{Illustration of the proposed hyperparameter detecting algorithm in RBFleX-NAS.}
\label{fig:ch2_self-detectingH}
\end{figure}

\subsection{Hyperparameter Detecting Algorithm (HDA)}
\label{sec:self-detectH}
The efficacy of the RBF kernel is largely subject to the value of hyperparameters such as $\gamma_k$, $\gamma_q$ in Equations (\ref{RBF_K}) and (3). Therefore, fine-tuning hyperparameters is crucial for the performance of the proposed algorithm. A large number of applications employ costly methods such as grid search \cite{grid1} to find the optimal hyperparameters. To minimize the cost, \cite{Gammaparameter} leverages the method proposed in \cite{fukunaga2013introduction} to set appropriate hyperparameters when deploying the RBF kernel in Support Vector Machine. The approach derives a projection vector on the basis of Fisher Linear Discriminant (FLD), which can generate an optimal regression line to separate the observed data samples into two non-overlapped classes. 
As our work aims to evaluate the similarity among $N$ input images, we develop an HDA inspired by FLD. Fig. \ref{fig:ch2_self-detectingH} illustrates the algorithm.
First, we select $M$ candidate networks and obtain the output matrix $\bf X$ and the input feature map matrix $\bf Y$ for each network using $N$ input images. Next, for any two vectors ${\bf v}_{i:}$ and ${\bf v}_{j:}$ in $\bf X$ or $\bf Y$, we define the mean value of the two vectors as
\begin{align}
    \label{mean_HDA_i}
    m_i = \frac{1}{L}\sum_{z=1}^L v_{iz} \\
    m_j = \frac{1}{L}\sum_{z=1}^L v_{jz}
\end{align}

\noindent where $L$ denotes the length of the vectors, $v_{iz}$ and $v_{jz}$ represent the $z$-th value in the selected vectors ${\bf v}_{i:}$ and ${\bf v}_{j:}$, respectively. We then define the squared error of the mean value of two vectors $D_{ij}$ as
\begin{align}
    \label{mean_HDA_j}
    D_{ij} = (m_i-m_j)^2
\end{align}

\begin{figure}[t!]
\includegraphics[width=0.48\textwidth]{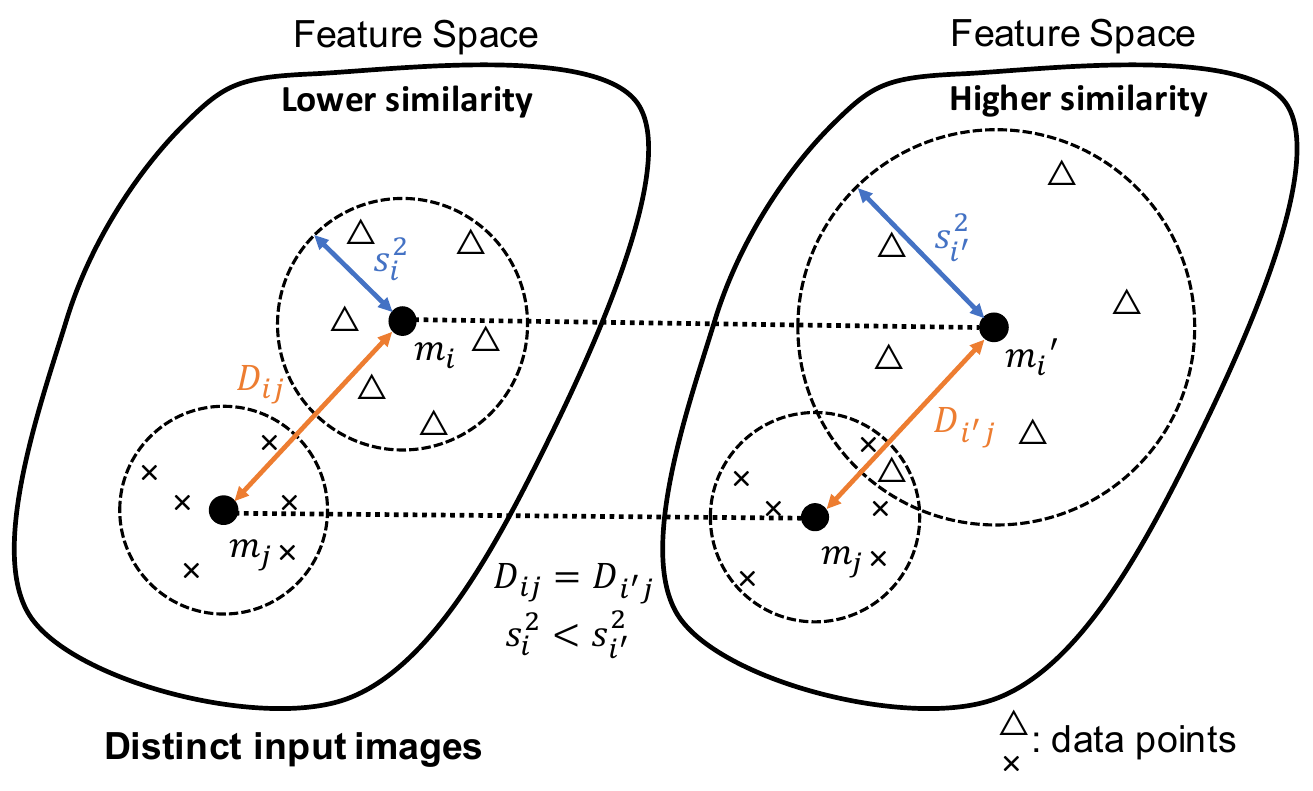}
\caption{Illustration of distinct input images that exhibit low similarity with small variance. $D_{ij}$ and $D_{i'j}$ are the squared error of the mean value between $m_i$ and $m_j$ as well as between $m_i'$ and $m_j$, respectively. $s_i$ and $s_i'$ are the variances of data points, respectively.
}
\label{fig:ch2_smallvar}
\end{figure}

\noindent However, evaluating the squared error between vectors solely by their mean values is not able to accurately assess their degree of overlap. As Fig.\ref{fig:ch2_smallvar} shows, the similarity between the vectors ${\bf v}_{i:}$ and ${\bf v}_{j:}$ from two entirely distinct input images is usually small and associated with much smaller variances (i.e., $s_i$ and $s_j$) than the squared error $D_{ij}$. If one of the variances (e.g., $s_i'$) is close to or larger than $D_{ij}$, the similarity will increase.
Therefore, we take the variance of the data samples in the two vectors into account. The variances of the two vectors are defined as
\begin{align}
    \label{var_HDA_i}
    s_i^2 = \frac{1}{L}\sum_{z=1}^L (v_{iz}-m_i)^2 \\
    s_j^2 = \frac{1}{L}\sum_{z=1}^L (v_{jz}-m_j)^2
\end{align}

\noindent We define $G_{ij}$ as a candidate hyperparameter for the RBF kernel, which is described as
\begin{align}
    \label{candi_gamma}
    G_{ij} = \frac{D_{ij}}{2(s_i^2+s_j^2)}\quad \text{if $D_{ij} \ne 0$}
\end{align}

\noindent Specifically, the candidate hyperparameter matrices ${\bf G}_k$ and ${\bf G}_q$ are defined by Equation (\ref{candi_gamma}), where the former is the hyperparameter matrix for the similarity of activation outputs and the latter is the matrix for the input feature maps of the last layer. 

\begin{figure}[t!]
\includegraphics[width=0.48\textwidth]{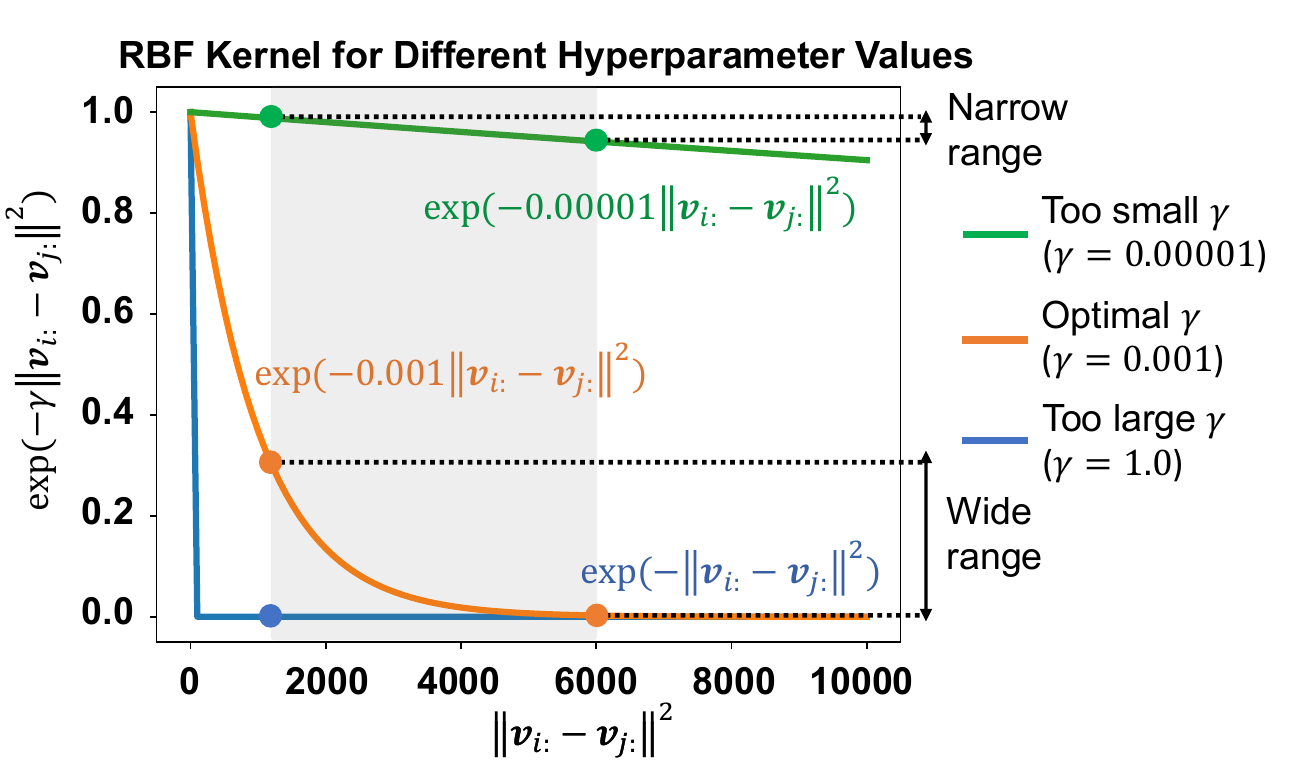}
\caption{RBF kernel for different hyperparameter values. The grey square indicates the range of Euclidean distance.}
\label{fig:ch2_diffHyperVal}
\end{figure}


Fig.\ref{fig:ch2_diffHyperVal} illustrates the RBF kernel function with different hyperparameter values, where the grey zone indicates the range of Euclidean distance $\|{\bf v}_{i:}-{\bf v}_{j:}\|$. A very large hyperparameter can make RBF kernel in Equation (2) and (3) insensitive to the difference in $\|{\bf v}_{i:}-{\bf v}_{j:}\|$, where ${\bf v}$ is ${\bf x}$ or ${\bf y}$.
Thus the hyperparameters, $\gamma_k$ and $\gamma_q$, should not be either too large or too small and an optimal hyperparameter is favorable to distinguish the similarity from various input images (i.e., with sufficient score variance).
In particular, the following equation 
\begin{align}
    \gamma = \frac{\log({1/\epsilon)}}{\max_{i, j}\|\mathbf{v_{i:}} - \mathbf{v_{j:}}\|^2}
\end{align}

\noindent guarantees that $\exp(-\gamma\|v_{i:} - v_{j:}\|^2) = \epsilon$ when $\|v_{i:} - v_{j:}\|^2$ becomes the largest 
and $\exp(-\gamma\|v_{i:} - v_{j:}\|^2) = \epsilon^{1/r} $ when $\|v_{i:} - v_{j:}\|^2$ becomes the smallest, 
where $r = \frac{\max_{i,j}|v_{i:} - v_{j:}\|^2}{\min_{i,j}|v_{i:} - v_{j:}\|^2}$ and we can choose an $\epsilon$ such that $[\epsilon, \epsilon^{1/r}]$ is as wide as possible. 
Therefore, in RBFleX-NAS, the fine-tuned hyperparameter $\gamma_k$ and $\gamma_q$ are obtained by the entry-wise minimal value in respective ${\bf G}_k$ and ${\bf G}_q$ as follows, where $\gamma_k >0$, $\gamma_q >0$ and $\min_{\substack{i,j}} {\bf G}(i,j)$ denotes the minimum entry in the matrix ${\bf G}$.
\begin{align}
    \label{gamma_k}
    \gamma_k = \min_{\substack{i,j}} {\bf G}_k(i,j) \text{ where } D_{ij}^{k} \neq 0 \\
    \gamma_q = \min_{\substack{i,j}} {\bf G}_q(i,j) \text{ where } D_{ij}^{q} \neq 0
\end{align}

\section{Experimental Result}
\label{sec:experiment}
We perform experiments to evaluate the efficacy of the proposed RBFleX-NAS in this section. Firstly, we investigate the effect of network initialization, minibatch size, and image batch on our score. Next, we compare the RBF kernel method with other similarity assessment methods. Thirdly, we evaluate the hyperparameter detecting algorithm. Subsequently, we compare RBFleX-NAS with gradient-based training-free NAS (i.e., grad\_norm\cite{abdelfattah2021zerocost}, snip\cite{lee2018snip}, synflow\cite{NEURIPS2020_46a4378f}, \textcolor{black}{ZiCo\cite{li2023zico}, and EProxy\cite{li2023extensible}}) and layer-based training-free NAS (i.e., NASWOT\cite{NASWOT}, \textcolor{black}{TE-NAS\cite{chen2021neural}}, and DAS\cite{DAS}) in terms of the correlation between the score and the network accuracy. Afterwards, we discuss the top-1 searched network in design spaces. Finally, we verify that RBFleX-NAS can score networks with various types of activation functions, leading to extended design space with activation function exploration.
\textcolor{black}{All computations in RBFleX-NAS are performed using 64-bit floating point precision to avoid the diminishing of similarity information when the associated hyperparameter value becomes too small.}
We utilize NAS-Bench-201\cite{dong2020nasbench201}, NATS-Bench\cite{NATS-Bench}, and Network Design Space (NDS)\cite{NDS} for the experiments on image classification workloads, \textcolor{black}{and TransNAS-Bench-101\cite{duan2021transnas} for object classification and semantic segmentation.} Specifically, NAS-Bench-201 consists of 15,625 networks pre-trained on CIFAR-10, CIFAR-100, and ImageNet. NATS-Bench integrates NATS-Bench-SSS for network size benchmarking and NATS-Bench-TSS for network topology exploration. NDS provides a variety of NAS search spaces, such as ResNet, Amoeba, DARTS, ENAS, PNAS, and NASNet. \textcolor{black}{TransNAS-Bench-101 provides 3,256 networks in macro-level search space with different depths, down-sampling layers, and enlarged-channel layers. It also incorporates 4,096 networks in micro-level (i.e., cell-level) search space with different operations adapting the feature map from node to node. These networks are pre-trained with the Taskonomy dataset\cite{zamir2018taskonomy}.} For the experiment on activation function design space, we create the Neural Network Activation Function Benchmark (NAFBee) based on VGG-19 over CIFAR-10 dataset and BERT over SST-2 dataset.
NAFBee is available at \textit{https://github.com/Edge-AI-Acceleration-Lab/NAFBee.git}

\begin{figure}
\centering
\includegraphics[width=0.48\textwidth]{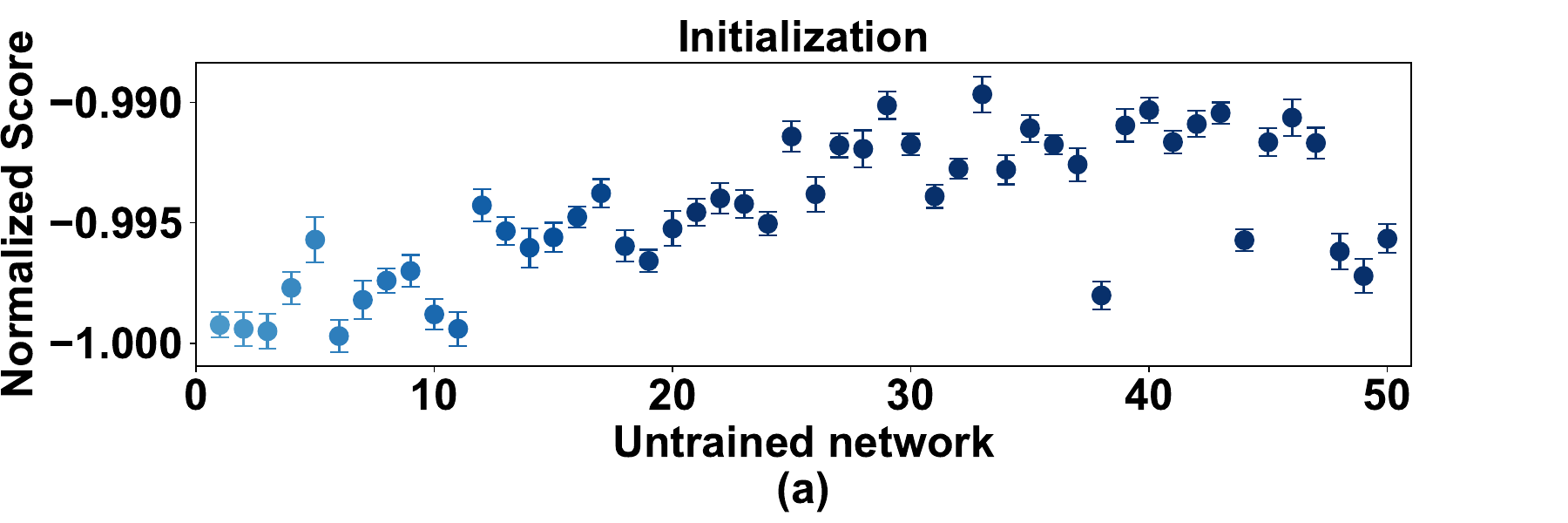}
\includegraphics[width=0.48\textwidth]{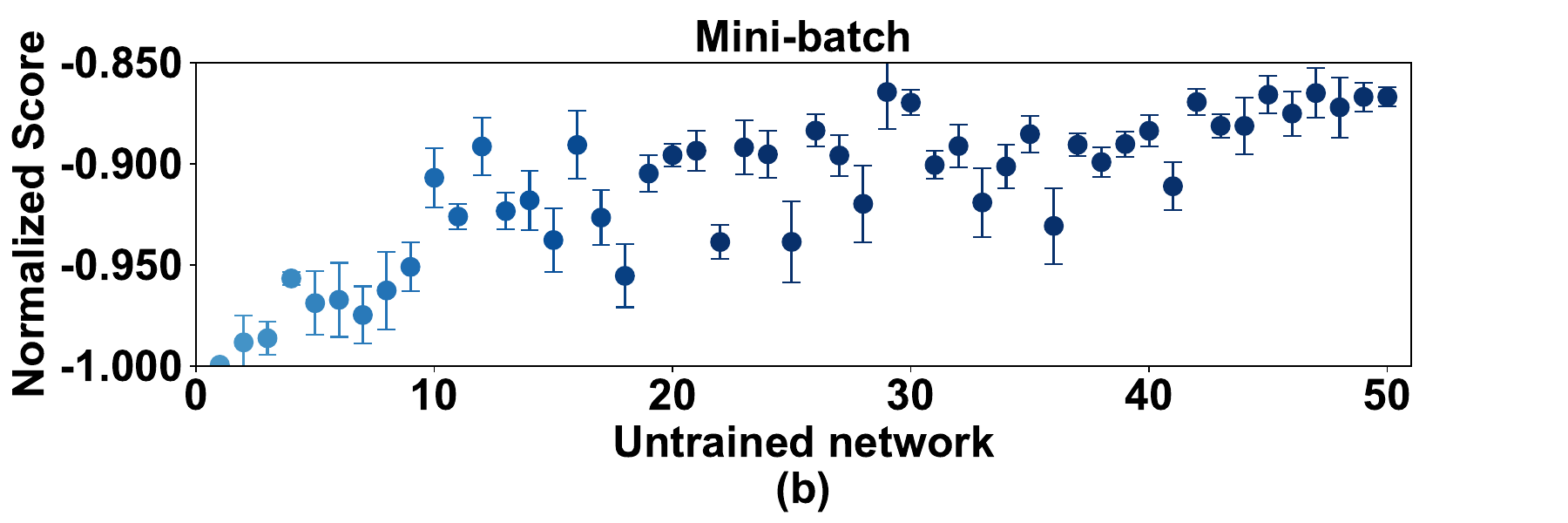}
\includegraphics[width=0.48\textwidth]{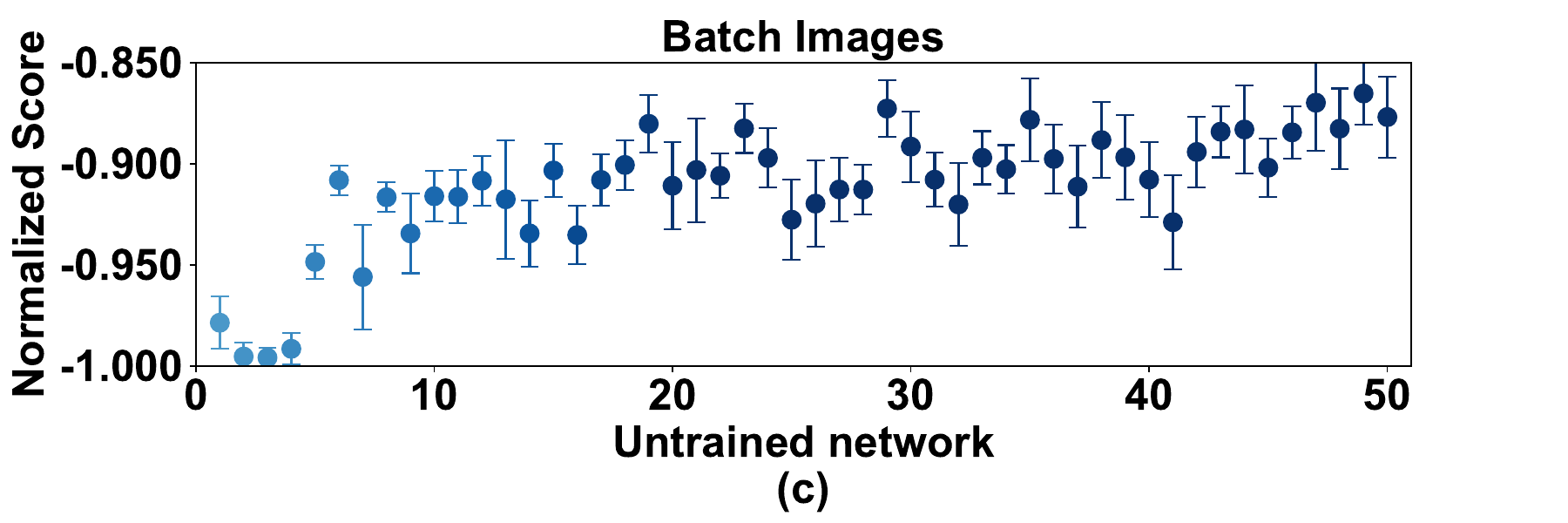}
\caption{Normalized network scores on NATS-Bench-SSS with respect to (a) network initialization, (b) minibatch size, and (c) different image batches. \textcolor{black}{50 networks are randomly sampled from NATS-Bench-SSS and ranked with its accuracy.
The first untrained network exhibits the
lowest accuracy, while the 50th network achieves the highest accuracy. The normalized score is obtained by dividing each network’s score by the minimum score derived from the first untrained network. As the RBFleX score is inherently negative, the normalized score starts at -1.0. A more negative normalized score corresponds to a better-performed network.}
}
\label{fig:ch2_plot}
\end{figure}

\subsection{\textcolor{black}{Impact of Network Initialization on RBFleX-NAS}}
\label{sec:initialization}
RBFleX-NAS initializes the weights of a network before scoring it. The network initialization can play a significant role in its final accuracy \cite{he2015delving, simonyan2015deep}. To investigate the effect of the initialization on the network performance, we randomly select 50 networks from a variety of network design space benchmarks, which are NATS-Bench-SSS \cite{NATS-Bench}, NAS-Bench-201\cite{dong2020nasbench201}, NDS(Amoeba)\cite{NDS}, NDS(DARTS), NDS(ENAS), and NDS(PNAS), respectively. We initialize their weights with random values and then score all selected networks 10 times with the ImageNet dataset. We investigate the effect of the network initialization via the score with min-max normalization on each trial.

Fig. \ref{fig:ch2_plot}(a) reveals the result from NATS-Bench-SSS. The accuracy of the untrained networks is exhibited in increasing order. Specifically, the first untrained network shows the lowest accuracy, while the 50th network achieves the highest accuracy. Apparently, RBFleX-NAS can consistently distinguish the networks in NATS-Bench-SSS with a negligible standard deviation of the score (i.e., standard deviation ranges from $\pm 4.1\times10^{-4}$ to $\pm 8.2\times10^{-4}$).
We also validate the network initialization in other design spaces and have observed a small standard deviation of 
$5.2\times10^{-3}$ for NAS-Bench-201, $1.6\times10^{-2}$ for NDS(Amoeba), $2.8\times10^{-3}$ for NDS(DARTS), $1.3\times10^{-2}$ for NDS(ENAS), and $3.4\times10^{-3}$ for NDS(PNAS), respectively. Therefore, our scoring approach is effective regardless of the initial weight values.

\subsection{\textcolor{black}{Impact of Minibatch Size on RBFleX-NAS}}
\label{sec:batch_size}

In the network scoring phase, RBFleX-NAS processes all the images in the same minibatch concurrently. We investigate the effect of the minibatch size by randomly selecting 50 networks from our design space benchmarks while ensuring distinct accuracy values. Subsequently, we score each network with a variety of batch sizes over the ImageNet dataset and normalize the scores to visualize the impact.

Fig. \ref{fig:ch2_plot}(b) shows the normalized score for minibatch size using NATS-bench-SSS. A larger normalized score indicates a better-performed network. 
RBFleX-NAS can successfully identify good architectures with a mean normalized score above -0.85 against different minibatch sizes. Tested with NAS-Bench-201 and NDS, RBFleX-NAS can evaluate the networks with very small standard deviations, which are less than $2.7\times10^{-2}$ in NAS-Bench-201, $8.1\times10^{-3}$ in NDS(Amoeba), $4.4\times10^{-3}$ in NDS(DARTS), $6.3\times10^{-3}$ in NDS(ENAS), and $4.9\times10^{-3}$ in NDS(PNAS). 
The tiny standard deviation indicates that RBFleX-NAS can consistently differentiate networks with distinct levels of accuracy in a design space, regardless of the minibatch size $N$.

\subsection{\textcolor{black}{Impact of Image Batch on RBFleX-NAS}}

\label{sec:batch_image}
RBFleX-NAS randomly selects $N$ images from a dataset to score a network. We thus analyze the effect of the image batch by selecting 50 networks from each benchmark (i.e., NATS-Bench-SSS, NAS-Bench-201, and NDS). RBFleX-NAS scores a network with 10 different batches on the ImageNet dataset.
We group 16 random images as a minibatch and score each network with 10 minibatches. We then investigate the effect of image batch with the same score normalization technique on each trial.

Fig. \ref{fig:ch2_plot}(c) illustrates the result of NATS-Bench-SSS with various image batches. The standard deviation of the score ranges from $\pm 4.6\times10^{-3}$ to $\pm 2.9\times10^{-2}$, which is significantly smaller than the score variance.
Besides, RBFleX-NAS can also evaluate the networks from NAS-Bench-201 and NDS with small standard deviations, which are less than $2.8\times10^{-2}$ for NAS-Bench-201, $1.4\times10^{-3}$ for NDS(Amoeba), $1.1\times10^{-3}$ for NDS(DARTS), $1.9\times10^{-3}$ for NDS(ENAS), and $1.8\times10^{-3}$ for NDS(PNAS), respectively. Therefore, RBFleX-NAS can effectively evaluate network architectures regardless of the image batch.

\subsection{Comparative Analysis of Similarity Assessment Methods}

RBFleX-NAS utilizes RBF kernels to assess the similarities among feature maps from a network. This experiment compares RBF kernel with other similarity assessment methods, such as linear kernel \cite{kung2014kernel}, the Peak Signal-to-Noise Ratio (PSNR) \cite{5596999}, and the Structural Similarity Index Measure (SSIM) \cite{1284395}. We randomly sample 1000 untrained networks from NATS-Bench-SSS and score those networks based on the four methods. Pearson and Kendall correlation coefficients are widely adopted to indicate the strength of the relationship between two variables \cite{cohen2009pearson}\cite{NASWOT}. We thus compare these similarity assessment methods by using mean Pearson and Kendall correlation coefficients over 10 sets of 1000 sampled networks. Our HDA derives the hyperparameters for RBF kernels. We score networks by CIFAR-10 dataset with a minibatch size of 16 and a number of networks for HDA of 10 (i.e., $N=16$, $M=10$).

\begin{figure}
\centering
\includegraphics[width=0.47\textwidth]{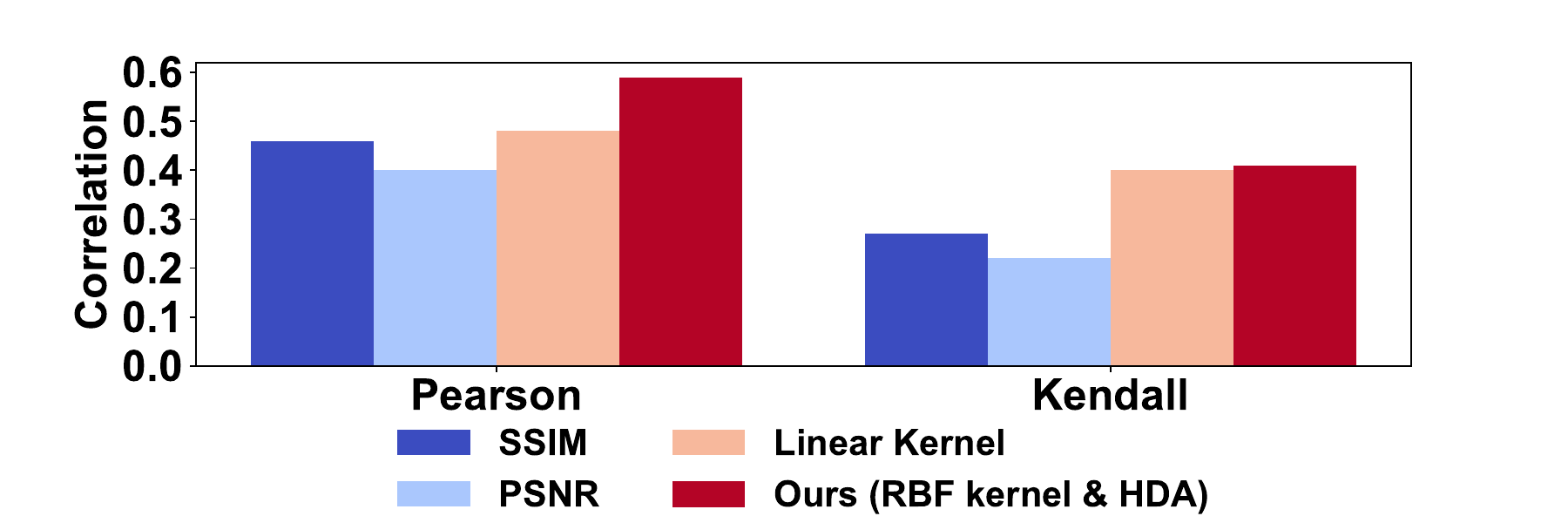} 
\caption{Pearson and Kendall Correlation of various similarity assessment methods. HDA detects the hyperparameter for RBF kernel.}
\label{fig:SAM}
\end{figure}

Fig. \ref{fig:SAM} reveals the RBF kernel method exhibits the highest mean Pearson correlation of 0.59 while the linear kernel method, PSNR, and SSIM yielded lower mean Pearson correlations of 0.48, 0.40 and 0.46, respectively. In terms of Kendall correlation, the RBF kernel method again achieves the highest mean Kendall correlation of 0.41. The linear kernel method, PSNR, and SSIM exhibit lower correlations of 0.40, 0.22, and 0.27, respectively. We confirm that the RBF kernel with HDA can effectively evaluate the similarity among feature maps.

\subsection{Evaluation of HDA}
 
To evaluate HDA, we randomly sample 1000 untrained networks from NATS-Bench-SSS and calculate the score of each network with the fine-tuned hyperparameters obtained from the algorithm.
We repeat this process 10 times over the CIFAR-10 dataset. To evaluate the performance of our HDA, we deploy Pearson correlation coefficient and Kendall correlation coefficient. Moreover, we compare our HDA with other two hyperparameter-detecting methods. Specifically, the first comparison method selects the maximal value in the candidate hyperparameter matrix ${\bf G}_k$ and ${\bf G}_q$, respectively while the second method randomly set values between $10^{-1}$ and $10^{-20}$. We consistently configure the minibatch size $N$ as 16, and the number of candidate networks $M$ as 10.

\begin{figure}[h!]
\centering
\includegraphics[width=0.45\textwidth]{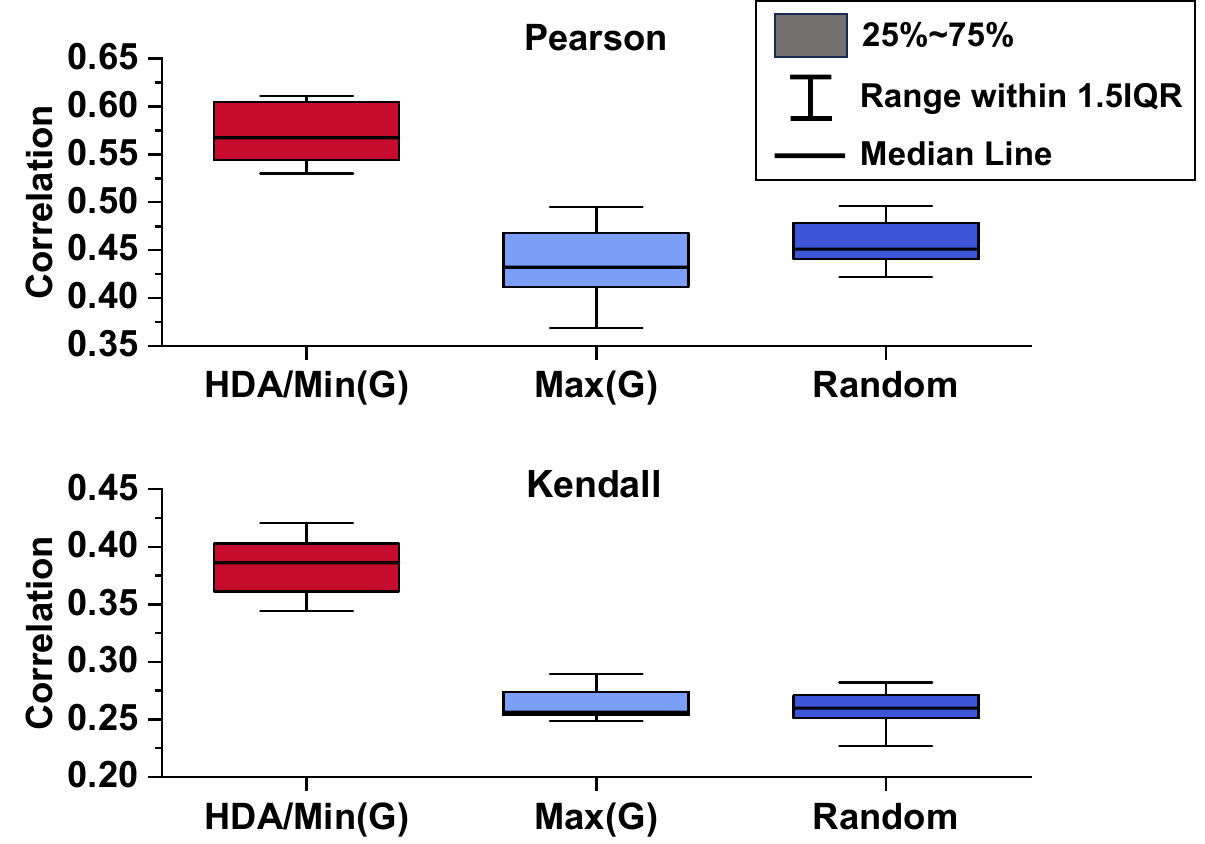} 
\caption{Pearson and Kendall correlations with hyperparameter detection algorithm (HDA) and random hyperparameters. The minibatch size is 16.}
\label{fig:ch2_HDA}
\end{figure}

Fig. \ref{fig:ch2_HDA} depicts the results of our HDA through a box-and-whisker plot, with mean Pearson and Kendall correlation values of 0.56 and 0.39, respectively. If selecting the maximal values in ${\bf G}_k$ and ${\bf G}_q$ for the hyperparameters as the first method, the mean correlations decrease to 0.43 and 0.26, respectively. Alternatively, if randomly setting the hyperparameters as the second method, the corresponding correlation values drop to 0.46 and 0.27, respectively. We conduct a $T-Test$ to assess the statistical difference between HDA and the maximum value as $T-Test$ is a statistical test widely adopted to compare the mean of two groups \cite{8808038}\cite{9060933}. $P-value$ calculated on $T-Test$ indicates the mean difference for two groups is statistically significant if $P-value$ is lower than 0.05. The results show a $P-value$ of $2.1\times10^{-5}$ for the Pearson correlation and $1.7\times10^{-10}$ for the Kendall correlation. Similarly, when comparing HDA with the random selection, the $T-Test$ yields a $P-value$ of $5.0\times10^{-4}$ for the Pearson correlation and $6.6\times10^{-7}$ for the Kendall correlation. As the $P-value$ is smaller than 0.05, we confirm that the proposed HDA is statistically significant to select better hyperparameters $\gamma_k$ and $\gamma_q$ for the RBF kernel.

\subsection{Evaluation of correlations between score and accuracy}
\textcolor{black}{In this experiment for image classification tasks, we compare RBFleX-NAS with eight NAS algorithms. Specifically, we benchmark it with gradient-based algorithms, including grad\_norm\cite{abdelfattah2021zerocost}, snip\cite{lee2018snip}, synflow\cite{NEURIPS2020_46a4378f}, EProxy\cite{li2023extensible}, and ZiCo\cite{li2023zico}. Simultaneously, we benchmark it with layer-based methods, including NASWOT\cite{NASWOT}, DAS\cite{DAS}, and TE-NAS\cite{chen2021neural}.} We test on various datasets, including CIFAR-10, CIFAR-100, and ImageNet, as well as with different network design spaces such as NAS-Bench-201, NATS-Bench-SSS, NDS(Amoeba), NDS(DARTS), NDS(ENAS), NDS(PNAS), NDS(ResNet), and NDS(NASNet). We configure the minibatch size $N$ as 16 and use 10 networks as the parameter $M$ to detect appropriate hyperparameters for RBF kernels. We also deploy the same two correlation coefficients and $T-test$ to verify the statistical significance. 

To compare the performance of RBFleX-NAS with the benchmarks, we conduct experiments on NAS-Bench-201, NATS-Bench-SSS, and NDS(ResNet), each with more than 10,000 networks. We randomly sample 1000 untrained networks and calculate the score of each network using RBFleX-NAS and the benchmarks. We repeat this process 10 times and compute the mean correlations for each algorithm. For design spaces with fewer networks (i.e., NDS(Amoeba), NDS(DARTS), NDS(ENAS), NDS(PNAS) and NDS(NASNet)), we deploy the NAS algorithms to evaluate in each design space.

\textcolor{black}{We further evaluate RBFleX-NAS on object classification and semantic segmentation tasks using TranNAS-Bench-101\cite{duan2021transnas} and Taskonomy dataset\cite{zamir2018taskonomy}. Specifically, we compare RBFleX-NAS with seven state-of-the-art training-free NAS algorithms, which are EProxy, TE-NAS, ZiCo, NASWOT, grad\_norm, snip, and synflow. RBFleX-NAS employs a minibatch size $N$ of 16 and utilizes 10 networks for HDA. All algorithms in the experiment score 3,256 networks in the macro-level design space and 4,096 networks in the micro-level design space. Pearson and Kendall correlations are calculated to evaluate the relationship between the scores and actual accuracies.}

\begin{table*}
\caption{Pearson and Kendall Correlation on NAS-Bench-201 and NATS-Bench-SSS}
\centering
\footnotesize
\begin{tabular}{l|ll|llllll}
\hline

\multirow{2}{3em}{Search Space} &\multicolumn{2}{l|}{Method} &\multicolumn{3}{l|}{Pearson Correlation} &\multicolumn{3}{l}{Kendall Correlation}\\
&Metric &Name &CIFAR-10 &CIFAR-100 &\multicolumn{1}{l|}{ImageNet} &CIFAR-10 &CIFAR-100 &ImageNet\\

\hline\hline

\multirow{8}{3em}{NAS\\Bench\\201} &\multirow{5}{11em}{Gradient-based\\(requires loss or labels)} 
               &grad{\_}norm
               &-0.009 &0.034 &0.087   
               &0.170 &0.204 &0.220 \\ 
              &&snip
               &0.003 &0.049 &0.033   
               &0.174 &0.212 &0.366 \\ 
              &&synflow
               &0.024 &0.050 &0.137   
               &0.296 &0.346 &0.224 \\ 
              &&\textcolor{black}{EProxy}
               &\textcolor{black}{-0.280} &\textcolor{black}{-0.878} &\textcolor{black}{-0.618}   
               &\textcolor{black}{-0.426} &\textcolor{black}{-0.528} &\textcolor{black}{-0.469}  \\ 
              &&\textcolor{black}{ZiCo}
               &\textcolor{black}{0.529} &\textcolor{black}{0.636} &\textcolor{black}{0.734}   
               &\textcolor{black}{0.564} &\textcolor{black}{0.603} &\textcolor{black}{0.583} \\ 
              
\cline{2-9}
              &\multirow{4}{11em}{Layer-based\\(no loss and label required)} 
              &DAS
               &0.717 &0.636 &0.582   
               &0.372 &0.406 &0.374 \\ 
              &&NASWOT
               &0.727 &0.638 &0.575   
               &0.374 &0.407 &0.376 \\ 
              &&\textcolor{black}{TE-NAS}
               &\textcolor{black}{0.051} &\textcolor{black}{0.095} &\textcolor{black}{0.266}   
               &\textcolor{black}{0.168} &\textcolor{black}{0.127} &\textcolor{black}{0.033} \\ 
              &&{\bf RBFleX-NAS}
               &\textcolor{black}{{\bf 0.898}} &\textcolor{black}{{\bf 0.723}} &{\bf 0.514}   
               &\textcolor{black}{{\bf 0.569}} &\textcolor{black}{{\bf 0.590}} &\textcolor{black}{{\bf 0.513}} \\ 

\hline

\multirow{8}{3em}{NATS\\Bench\\SSS} &\multirow{5}{11em}{Gradient-based\\(requires loss or labels)} 
               &grad{\_}norm
               &0.606 &0.437 &0.494   
               &0.426 &0.317 &0.376 \\ 
              &&snip
               &0.698 &0.531 &0.669   
               &0.511 &0.394 &0.523 \\ 
              &&synflow
               &0.508 &0.453 &0.560   
               &0.718 &0.579 &0.771 \\ 
              &&\textcolor{black}{EProxy}
               &\textcolor{black}{-0.346} &\textcolor{black}{0.376} &\textcolor{black}{0.166}   
               &\textcolor{black}{-0.237} &\textcolor{black}{0.279} &\textcolor{black}{0.130}  \\ 
              &&\textcolor{black}{ZiCo}
               &\textcolor{black}{0.907} &\textcolor{black}{0.738} &\textcolor{black}{0.871}   
               &\textcolor{black}{0.729} &\textcolor{black}{0.555} &\textcolor{black}{0.697} \\ 
              
\cline{2-9}
              &\multirow{4}{11em}{Layer-based\\(no loss and labels required)} 
              &DAS
               &0.484 &0.241 &0.523   
               &0.318 &0.159 &0.366 \\ 
              &&NASWOT
               &0.484 &0.241 &0.524   
               &0.318 &0.160 &0.367 \\ 
              &&\textcolor{black}{TE-NAS}
               &\textcolor{black}{-0.146} &\textcolor{black}{-0.151} &\textcolor{black}{-0.363}   
               &\textcolor{black}{-0.114} &\textcolor{black}{-0.122} &\textcolor{black}{-0.313} \\ 
              &&{\bf RBFleX-NAS}
               &\textcolor{black}{{\bf 0.607}} &\textcolor{black}{{\bf 0.855}} &\textcolor{black}{{\bf 0.869}}   
               &\textcolor{black}{{\bf 0.419}} &\textcolor{black}{{\bf 0.639}} &\textcolor{black}{{\bf 0.649}} \\ 

\hline

\end{tabular} 
\label{table:201SSScorr}
\end{table*}

\begin{figure}
\centering
\includegraphics[width=0.48\textwidth]{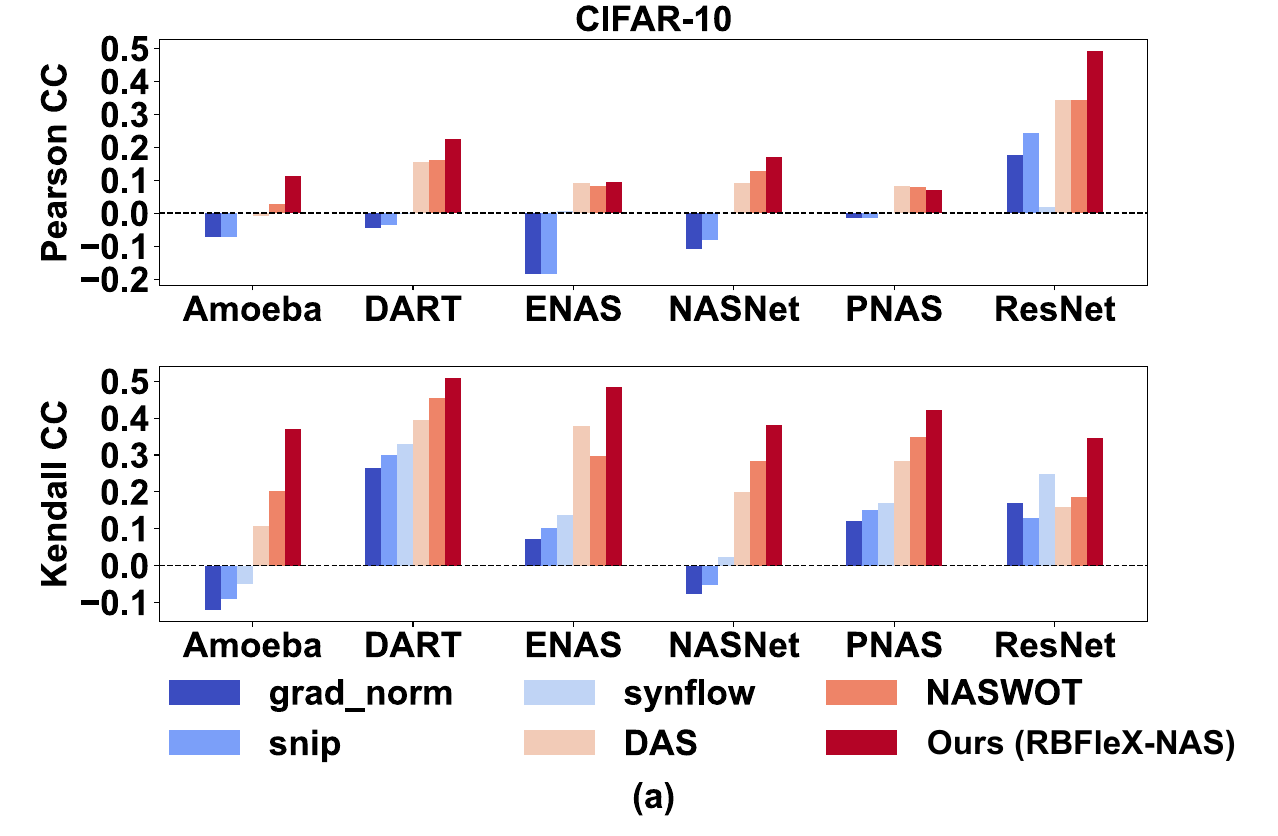}
\includegraphics[width=0.48\textwidth]{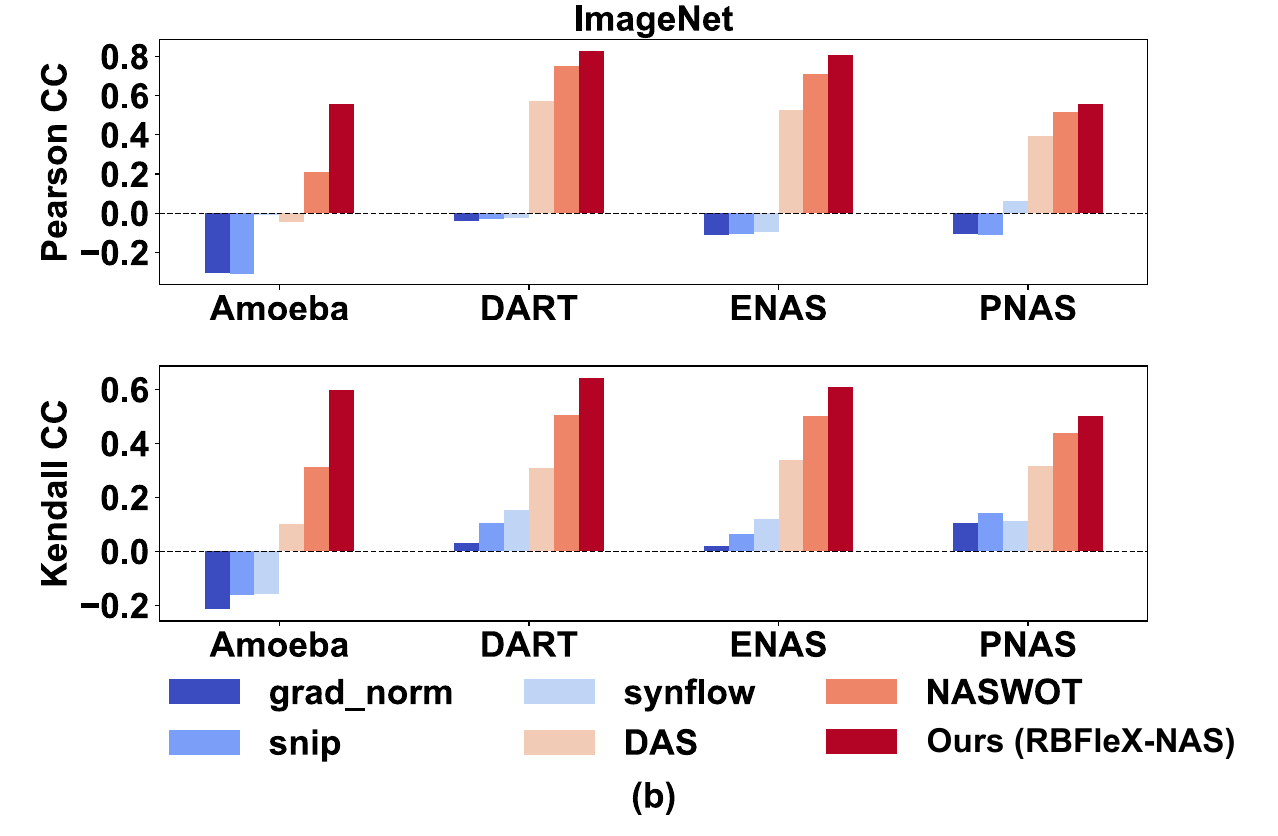}
\caption{Pearson and Kendall Correlations on NDS benchmark. (a) over CIAFR-10. (b) over ImageNet}
\label{fig:NDSbench}
\end{figure}

Table \ref{table:201SSScorr} shows the experiment results on NAS-Bench-201 and NATS-Bench-SSS over CIFAR-10, CIFAR-100, and ImageNet. Fig. \ref{fig:NDSbench} shows the Pearson and Kendall correlation on NDS benchmark over CIFAR-10 and ImageNet. \textcolor{black}{Though RBFleX-NAS exhibits lower correlation than ZiCo in certain cases, it can identify networks with higher accuracy using less search time, which will be elaborated in Section \ref{Analysis_of_Architecture_Search}. Moreover,
it shows higher Kendall correlations compared to all layer-based training-free NAS methods, including DAS, NASWOT and TE-NAS across both search spaces.}
\textcolor{black}{For NATS-Bench-SSS, our RBFleX-NAS also outperforms layer-based training-free algorithms (i.e., NASWOT, DAS, and TE-NAS) in terms of Pearson correlation}. Fig. \ref{fig:NDSbench}(a) illustrates that RBFleX-NAS consistently outperforms state-of-the-art layer-based and gradient-based training-free NAS methods in NDS over CIFAR-10, as it achieves higher positive correlations across all design spaces. In contrast, the gradient-based proxies which are evaluated on NDS(Amoeba) and NDS(NASNet) show negative Kendall correlations.

\begin{figure}
\centering
\includegraphics[width=0.48\textwidth]{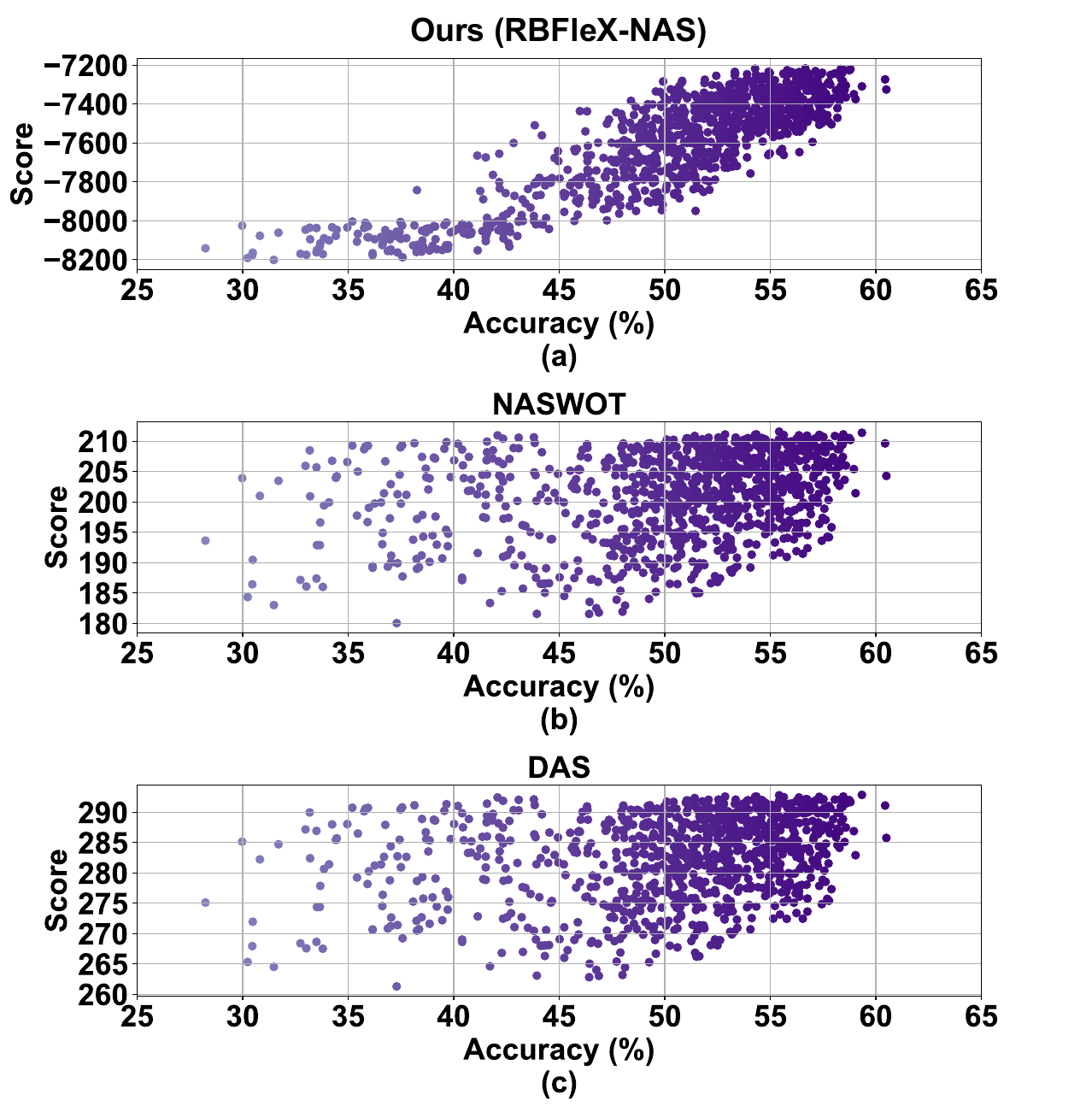}
\caption{Score vs final accuracy over CIFAR-100 in NATS-Bench-SSS with a minibatch size of 16. (a) Score by RBFleX-NAS. (b) Score by NASWOT. (c) Score by DAS.}
\label{fig:ch2_201}
\end{figure}

In terms of the Pearson and Kendall correlation over CIFAR-100, our RBFleX-NAS outperforms other layer-based and gradient-based algorithms significantly in NATS-Bench-SSS, achieving a higher Pearson correlation of 0.855 and a higher Kendall correlation of 0.639. The experiment on NATS-Bench-SSS over ImageNet shows that RBFleX-NAS demonstrates apparently higher correlations compared to other layer-based traning-free NAS frameworks. Specifically, the Pearson and Kendall correlations are 0.869 and 0.649, respectively, despite the increased challenge posed by the ImageNet dataset.

Fig. \ref{fig:NDSbench}(b) shows RBFleX-NAS also achieves higher correlation for all design space from NDS while grad\_norm, snip, and synflow proxies achieve negative Kendall correlation for NDS(Amoeba). From the data in the tables, we observe that RBFleX-NAS can achieve a positive correlation between the score and the final accuracy of the networks. Moreover, we highlight the better results from RBFlex-NAS, which reveals that RBFleX-NAS outperforms NASWOT and DAS in predicting the final trained accuracy of the networks. Besides, RBFleX-NAS demonstrates positive correlations across all design spaces, while the gradient-based proxies yield negative correlations for NDS design spaces, indicating RBFleX-NAS's versatility.

\begin{figure}
\centering
\includegraphics[width=0.48\textwidth]{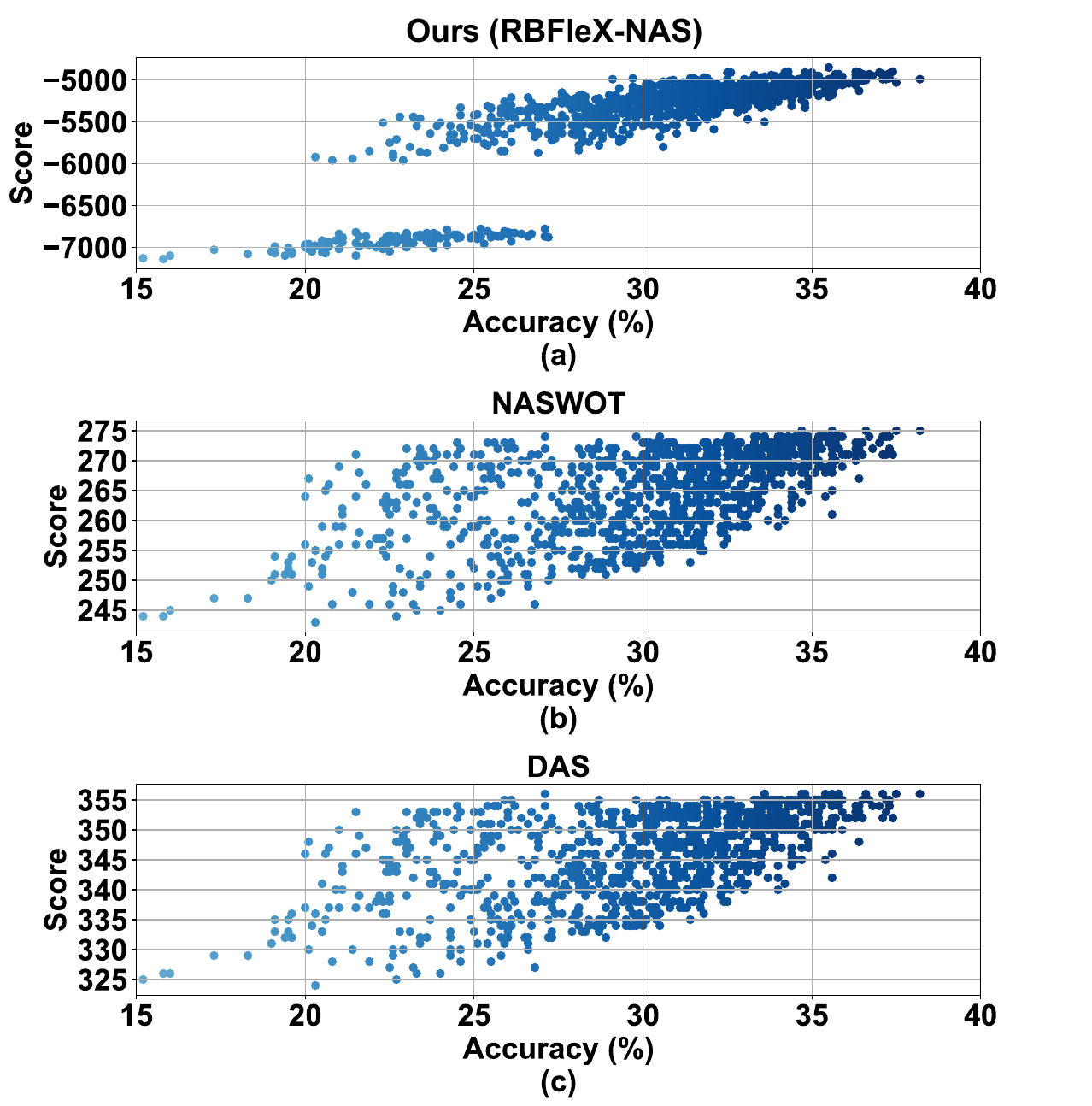}
\caption{Score vs final accuracy over ImageNet in NATS-Bench-SSS with a minibatch size of 16. (a) Score by RBFleX-NAS. (b) Score by NASWOT. (c) Score by DAS.}
\label{fig:ch2_SSS}
\end{figure}

Fig. \ref{fig:ch2_201} shows the scatter plots between the score and the validated accuracy of the networks over CIFAR-100 in NATS-Bench-SSS by RBFleX-NAS, NASWOT, and DAS, respectively. RBFleX-NAS successfully evaluates the under-performing networks with an accuracy below 40\% and effectively distinguishes them from other well-performed candidates. RBFleX-NAS also demonstrates a more linear dependency between the score and the accuracy of the networks compared to NASWOT and DAS.

Fig. \ref{fig:ch2_SSS} exhibits the dependency between the score and the final accuracy of the networks from NATS-Bench-SSS over the ImageNet dataset. The plot shows that RBFleX-NAS outperforms NASWOT and DAS in identifying better-performed networks. Specifically, RBFleX-NAS is able to distinguish better networks (i.e., accuracy higher than 30\%) from inferior networks (i.e., accuracy between 15\%-25\%) while NASWOT and DAS can assign almost the same score to the networks with a lower accuracy (i.e., between 20\% and 30\%) and the networks with significantly higher accuracy (i.e., 30\% or above).

\textcolor{black}{Table \ref{table:Tran101corr} shows Pearson and Kendall correlations on object classification and semantic segmentation tasks using TransNAS-Bench-101. On macro-level object classification (OC Macro), RBFleX-NAS achieves a higher accuracy of 45.42\% with higher Pearson and Kendall correlations. On micro-level object classification (OC Micro), RBFleX-NAS also exhibits a higher accuracy of 43.26\% with a higher Kendall correlation compared to TE-NAS and NASWOT. For macro-level semantic segmentation (SS Macro), the Pearson and Kendall correlations of RBFleX-NAS are 0.853 and 0.668 with the highest mIoU of 29.15\%. For micro-level semantic segmentation (SS Micro), RBFleX-NAS achieves a Pearson correlation of 0.547  and a Kendall correlation of 0.381 with the highest mIoU of 25.33\%. This illustrates that RBFleX-NAS is capable of identifying higher-accuracy networks with good correlation in object classification and semantic segmentation.}

\begin{table}
\caption{\textcolor{black}{Pearson and Kendall Correlation for Object Classification (OC) and Semantic Segmentation (SS) Tasks using TransNAS-Bench-101.}}
\centering
\footnotesize
\begin{tabular}{l|ll|ccc}
\hline

\multirow{2}{3em}{Search Space} &\multicolumn{2}{l|}{Method} &\multicolumn{2}{l|}{Correlation} &\\
&Metric &Name &\multicolumn{1}{c}{Pearson}  &\multicolumn{1}{c|}{Kendall} &\multicolumn{1}{c}{Top-1 (\%)}\\

\hline\hline

\multirow{8}{3em}{OC\\Macro} &\multirow{5}{3em}{Gradient-based} 
               &grad{\_}norm
               &-0.291   
               &-0.194
               &44.43\\ 
              &&snip
               &0.073   
               &0.009
               &44.49\\ 
              &&synflow
               &-0.788 
               &-0.586
               &42.42 \\ 
              &&EProxy
               &-0.026 
               &-0.049
               &44.28\\ 
              &&ZiCo
               &-0.167 
               &-0.110
               &43.45\\ 
              
\cline{2-6}
              &\multirow{3}{3em}{Layer-based}
              &TE-NAS
               &0.011 
               &-0.029
               &45.02\\ 
              &&NASWOT
               &0.734   
               &0.554
               &45.42\\ 
              &&{\bf RBFleX-NAS}
               &{\bf 0.740}    
               &{\bf 0.555}
               &{\bf 45.42}\\ 

\hline

\multirow{8}{3em}{OC\\Micro} &\multirow{5}{3em}{Gradient-based} 
               &grad{\_}norm
               &0.130   
               &0.094
               &38.40\\ 
              &&snip
               &0.068   
               &0.407
               &42.26\\ 
              &&synflow
               &0.035   
               &0.102
               &26.55\\ 
              &&EProxy
               &-0.008   
               &-0.019
               &40.15\\ 
              &&ZiCo
               &0.632   
               &0.398
               &43.05\\ 
              
\cline{2-6}
              &\multirow{3}{3em}{Layer-based}
              &TE-NAS
               &-0.030   
               &-0.045
               &42.57\\ 
              &&NASWOT
               &0.598   
               &0.322
               &27.93\\ 
              &&{\bf RBFleX-NAS}
               &{\bf 0.461}   
               &{\bf 0.338}
               &{\bf 43.26}\\ 

\hline

\multirow{2}{3em}{Search Space} &\multicolumn{2}{l|}{Method} &\multicolumn{2}{l|}{Correlation} &\\
&Metric &Name &\multicolumn{1}{c}{Pearson}  &\multicolumn{1}{c|}{Kendall} &\multicolumn{1}{c}{mIoU (\%)}\\

\hline\hline

\multirow{8}{3em}{SS\\Macro} &\multirow{5}{3em}{Gradient-based} 
               &grad{\_}norm
               &-0.125   
               &-0.083
               &28.07\\ 
              &&snip
               &0.059   
               &0.029
               &28.57\\ 
              &&synflow
               &-0.765   
               &-0.57
               &21.05\\ 
              &&EProxy
               &-0.049   
               &-0.062
               &21.18\\ 
              &&ZiCo
               &-0.087   
               &-0.063
               &26.79\\ 
              
\cline{2-6}
              &\multirow{3}{3em}{Layer-based}
              &TE-NAS
               &-0.002   
               &0.001
               &24.85\\ 
              &&NASWOT
               &0.854   
               &0.674
               &29.15\\ 
              &&{\bf RBFleX-NAS}
               &{\bf 0.853}    
               &{\bf 0.668}
               &{\bf 29.15}\\ 

\hline

\multirow{8}{3em}{SS\\Micro} &\multirow{5}{3em}{Gradient-based} 
               &grad{\_}norm
               &0.216   
               &0.385
               &22.68\\ 
              &&snip
               &0.098   
               &0.488
               &24.43\\ 
              &&synflow
               &0.035   
               &0.230
               &5.41\\ 
              &&EProxy
               &-0.036   
               &-0.037
               &15.25\\ 
              &&ZiCo
               &0.620   
               &0.416
               &24.43\\ 
              
\cline{2-6}
              &\multirow{3}{3em}{Layer-based}
              &TE-NAS
               &-0.015   
               &-0.115
               &9.34\\ 
              &&NASWOT
               &0.592   
               &0.339
               &21.87\\ 
              &&{\bf RBFleX-NAS}
               &{\bf 0.547}    
               &{\bf 0.381}
               &{\bf 25.33}\\ 

\hline

\end{tabular} 
\label{table:Tran101corr}
\end{table}

\begin{table*}
\caption{Comparison on architecture search. Weight Sharing Method Shares Weights to Allow for Joint Training with Candidate Networks. \textcolor{black}{In NAS-Bench-201, the top-performing network for each dataset is identified using the CIFAR-10 dataset. In NATS-Bench-SSS, the top-performing network for each dataset is determined using its respective dataset.} 
}
\centering
\footnotesize
\begin{tabular}{l|ll|llllll}
\hline

\multirow{2}{3em}{Search Space} &\multicolumn{2}{l|}{Method} 
&CIFAR-10 &\multicolumn{1}{l|}{Search} &CIFAR-100 &\multicolumn{1}{l|}{Search} &ImageNet &Search\\
&Metric &Name &Accuracy (\%) &\multicolumn{1}{l|}{Cost (s)} &Accuracy (\%) &\multicolumn{1}{l|}{Cost (s)} &Accuracy (\%) &Cost (s) \\

\hline\hline

\multirow{31}{2em}{NAS\\Bench\\201} &\multirow{6}{11em}{Weight Sharing} 
              &RSPS
              &88.10$\pm$1.06 &\textcolor{black}{1524} 
              &66.31$\pm$1.29 &\textcolor{black}{1524}  
              &38.21$\pm$2.11 &\textcolor{black}{1524} \\
              
              &&DARTS(1st)                                 
              &59.02$\pm$13.02 &\textcolor{black}{9631} 
              &14.99$\pm$0.05 &\textcolor{black}{9631} 
              &16.43$\pm$0.79 &\textcolor{black}{9631} \\
              
              &&DARTS(2nd)                                  
              &39.77$\pm$0.12 &\textcolor{black}{28457}
              &15.03$\pm$0.07 &\textcolor{black}{28457}
              &15.93$\pm$0.69 &\textcolor{black}{28457}\\
              
              &&GDAS                                      
              &89.14$\pm$1.81 &\textcolor{black}{4506}
              &70.65$\pm$1.51 &\textcolor{black}{4506}
              &41.48$\pm$3.55 &\textcolor{black}{4506}\\
              
              &&ENAS                                  
              &90.17$\pm$0.23 &\textcolor{black}{2353} 
              &70.76$\pm$0.69 &\textcolor{black}{2353}  
              &40.68$\pm$0.60 &\textcolor{black}{2353} \\
              
              &&SETN                                       
              &88.78$\pm$1.10 &\textcolor{black}{5886} 
              &66.41$\pm$1.51 &\textcolor{black}{5886} 
              &38.96$\pm$2.43 &\textcolor{black}{5886}\\
              
\cline{2-9}
              &\multirow{15}{11em}{Gradient-based\\Training-free\\(requires loss or labels)} 
              &\textcolor{black}{ZiCo(S=10)} 
              &\textcolor{black}{89.31$\pm$1.09} &\textcolor{black}{1.28}
              &\textcolor{black}{68.87$\pm$1.84} &\textcolor{black}{1.28}
              &\textcolor{black}{41.83$\pm$3.68} &\textcolor{black}{1.28}\\
              
              &&\textcolor{black}{ZiCo(S=100)} 
              &\textcolor{black}{89.89$\pm$0.56} &\textcolor{black}{12.82}
              &\textcolor{black}{70.13$\pm$1.34} &\textcolor{black}{12.82}
              &\textcolor{black}{42.99$\pm$3.68} &\textcolor{black}{12.82}\\

              &&\textcolor{black}{ZiCo(S=1000)}         
              &\textcolor{black}{90.18$\pm$0.25} &\textcolor{black}{128.24}
              &\textcolor{black}{70.28$\pm$1.12} &\textcolor{black}{128.24}
              &\textcolor{black}{43.09$\pm$1.95} &\textcolor{black}{128.24} \\

              &&\textcolor{black}{EProxy(S=10)}
              &\textcolor{black}{76.65$\pm$10.69} &\textcolor{black}{16.12}
              &\textcolor{black}{50.47$\pm$12.09} &\textcolor{black}{16.12}
              &\textcolor{black}{25.21$\pm$8.77} &\textcolor{black}{16.12} \\
              
              &&\textcolor{black}{EProxy(S=100)}
              &\textcolor{black}{54.70$\pm$15.79} &\textcolor{black}{179.56}
              &\textcolor{black}{28.57$\pm$14.79} &\textcolor{black}{179.56}
              &\textcolor{black}{15.53$\pm$4.69} &\textcolor{black}{179.56}\\

              &&\textcolor{black}{EProxy(S=1000)}         
              &\textcolor{black}{53.51$\pm$13.42} &\textcolor{black}{1760}
              &\textcolor{black}{27.14$\pm$11.83} &\textcolor{black}{1760}
              &\textcolor{black}{15.07$\pm$1.89} &\textcolor{black}{1760} \\

              &&\textcolor{black}{Synflow(S=10)} 
              &\textcolor{black}{86.07$\pm$6.18} &\textcolor{black}{0.85}
              &\textcolor{black}{62.66$\pm$7.55} &\textcolor{black}{0.85}
              &\textcolor{black}{34.94$\pm$6.18} &\textcolor{black}{0.85}\\
              
              &&\textcolor{black}{Synflow(S=100)} 
              &\textcolor{black}{19.12$\pm$25.49} &\textcolor{black}{8.81}
              &\textcolor{black}{8.91$\pm$21.46} &\textcolor{black}{8.81}
              &\textcolor{black}{5.57$\pm$3.43} &\textcolor{black}{8.81}\\

              &&\textcolor{black}{Synflow(S=1000)}         
              &\textcolor{black}{9.71$\pm$1.45} &\textcolor{black}{90.65}
              &\textcolor{black}{0.99$\pm$2.99} &\textcolor{black}{90.65}
              &\textcolor{black}{0.83$\pm$6.85} &\textcolor{black}{90.65} \\

              &&\textcolor{black}{grad\_norm(S=10)} 
              &\textcolor{black}{86.68$\pm$5.37} &\textcolor{black}{0.78}
              &\textcolor{black}{64.23$\pm$6.94} &\textcolor{black}{0.78}
              &\textcolor{black}{35.35$\pm$7.57} &\textcolor{black}{0.78}\\
              
              &&\textcolor{black}{grad\_norm(S=100)}
              &\textcolor{black}{87.59$\pm$1.79} &\textcolor{black}{8.12}
              &\textcolor{black}{64.56$\pm$3.99} &\textcolor{black}{8.12}
              &\textcolor{black}{33.48$\pm$7.65} &\textcolor{black}{8.12}\\

              &&\textcolor{black}{grad\_norm(S=1000)}         
              &\textcolor{black}{86.37$\pm$1.66} &\textcolor{black}{80.97}
              &\textcolor{black}{61.91$\pm$3.45} &\textcolor{black}{80.97}
              &\textcolor{black}{27.46$\pm$9.25} &\textcolor{black}{80.97} \\

              &&\textcolor{black}{Snip(S=10)} 
              &\textcolor{black}{89.27$\pm$1.42} &\textcolor{black}{8.94}
              &\textcolor{black}{68.71$\pm$2.40} &\textcolor{black}{8.94}
              &\textcolor{black}{40.98$\pm$4.19} &\textcolor{black}{8.94}\\
              
              &&\textcolor{black}{Snip(S=100)}
              &\textcolor{black}{89.82$\pm$1.14} &\textcolor{black}{90.13}
              &\textcolor{black}{69.92$\pm$2.02} &\textcolor{black}{90.13}
              &\textcolor{black}{41.32$\pm$4.94} &\textcolor{black}{90.13}\\

              &&\textcolor{black}{Snip(S=1000)}         
              &\textcolor{black}{89.70$\pm$0.82} &\textcolor{black}{912.86}
              &\textcolor{black}{67.69$\pm$2.68} &\textcolor{black}{912.86}
              &\textcolor{black}{40.59$\pm$4.57} &\textcolor{black}{912.86}\\
\cline{2-9}
              &\multirow{10}{11em}{Layer-based\\Training-free\\(no loss and label required)} 
              &Random                    
              &83.94$\pm$12.60 &N/A 
              &61.62$\pm$11.82 &N/A 
              &33.78$\pm$8.94 &N/A\\
              
              &&NASWOT(S=10)              
              &89.20$\pm$1.11 &\textcolor{black}{0.64} 
              &68.70$\pm$1.89 &\textcolor{black}{0.64}  
              &41.51$\pm$3.24 &\textcolor{black}{0.64} \\
              
              &&NASWOT(S=100)             
              &89.55$\pm$0.83 &\textcolor{black}{7.15} 
              &69.45$\pm$1.69 &\textcolor{black}{7.15}  
              &43.04$\pm$2.58 &\textcolor{black}{7.15} \\

              &&\textcolor{black}{NASWOT(S=1000)}        
              &\textcolor{black}{89.70$\pm$0.84} &\textcolor{black}{84.31}
              &\textcolor{black}{69.65$\pm$1.45} &\textcolor{black}{84.31}
              &\textcolor{black}{43.61$\pm$2.52} &\textcolor{black}{84.31} \\
              
              &&\textcolor{black}{TE-NAS(S=10)} 
              &\textcolor{black}{88.28$\pm$1.52} &\textcolor{black}{10.43} 
              &\textcolor{black}{66.39$\pm$3.04} &\textcolor{black}{10.43}
              &\textcolor{black}{39.02$\pm$4.04} &\textcolor{black}{10.43}\\
        
              &&\textcolor{black}{TE-NAS(S=100)}
              &\textcolor{black}{89.60$\pm$0.38} &\textcolor{black}{106.38}
              &\textcolor{black}{67.68$\pm$0.56} &\textcolor{black}{106.38}
              &\textcolor{black}{41.86$\pm$1.86} &\textcolor{black}{106.38}\\

              &&\textcolor{black}{TE-NAS(S=1000)}         
              &\textcolor{black}{89.76$\pm$0.23} &\textcolor{black}{1052}
              &\textcolor{black}{67.83$\pm$0.55} &\textcolor{black}{1052}
              &\textcolor{black}{42.61$\pm$1.37} &\textcolor{black}{1052} \\
              
              &&{\bf RBFleX-NAS(S=10)}          
              &\textcolor{black}{{\bf 93.23$\pm$1.29}} &\textcolor{black}{{\bf 0.89}} 
              &\textcolor{black}{{\bf 70.53$\pm$1.70}} &\textcolor{black}{{\bf 0.89}} 
              &\textcolor{black}{{\bf 41.54$\pm$2.65}} &\textcolor{black}{{\bf 0.89}}\\

              &&{\bf RBFleX-NAS(S=100)}         
              &\textcolor{black}{{\bf 93.36$\pm$0.94}} &\textcolor{black}{{\bf 9.17}} 
              &\textcolor{black}{{\bf 70.90$\pm$0.93}} &\textcolor{black}{{\bf 9.17}} 
              &\textcolor{black}{{\bf 45.81$\pm$0.57}} &\textcolor{black}{{\bf 9.17}}\\

              &&{\bf RBFleX-NAS(S=1000)}         
              &\textcolor{black}{{\bf 93.30$\pm$0.35}} &\textcolor{black}{{\bf 95.36}} 
              &\textcolor{black}{{\bf 70.99$\pm$0.21}} &\textcolor{black}{{\bf 95.36}} 
              &\textcolor{black}{{\bf 44.57$\pm$0.70}} &\textcolor{black}{{\bf 95.36}}\\

\hline

\hline

\multirow{28}{2em}{NATS\\Bench\\SSS} &\multirow{3}{11em}{Weight Sharing} 
              &TAS 
              &93.40$\pm$0.00 &\textcolor{black}{1946}
              &70.72$\pm$0.00 &\textcolor{black}{3863} 
              &47.17$\pm$0.00 &\textcolor{black}{7776} \\
              
              &&FBNet-v2                                 
              &93.04$\pm$0.18 &\textcolor{black}{1377}
              &69.96$\pm$0.69 &\textcolor{black}{2733} 
              &45.05$\pm$0.56 &\textcolor{black}{5053}\\

              &&\textcolor{black}{TuNAS}
              &\textcolor{black}{85.84$\pm$0.25} &\textcolor{black}{1315}
              &\textcolor{black}{57.84$\pm$4.19} &\textcolor{black}{2672} 
              &\textcolor{black}{36.06$\pm$0.10} &\textcolor{black}{6228}\\
\cline{2-9}
              &\multirow{15}{11em}{Gradient-based\\Training-free\\(requires loss or labels)} 
              &\textcolor{black}{ZiCo(S=10)}
              &\textcolor{black}{89.39$\pm$0.50} &\textcolor{black}{0.71}
              &\textcolor{black}{67.46$\pm$1.83} &\textcolor{black}{0.77}
              &\textcolor{black}{42.50$\pm$1.83} &\textcolor{black}{13.90} \\
              
              &&\textcolor{black}{ZiCo(S=100)}
              &\textcolor{black}{89.83$\pm$0.32} &\textcolor{black}{7.07}
              &\textcolor{black}{68.74$\pm$0.97} &\textcolor{black}{7.25}
              &\textcolor{black}{44.22$\pm$0.96} &\textcolor{black}{135.93} \\

              &&\textcolor{black}{ZiCo(S=1000)}         
              &\textcolor{black}{90.03$\pm$0.26} &\textcolor{black}{69.90}
              &\textcolor{black}{69.36$\pm$0.60} &\textcolor{black}{69.76}
              &\textcolor{black}{44.90$\pm$0.79} &\textcolor{black}{1327} \\

              &&\textcolor{black}{EProxy(S=10)}
              &\textcolor{black}{87.05$\pm$1.38} &\textcolor{black}{9.94}
              &\textcolor{black}{65.50$\pm$2.67} &\textcolor{black}{9.10}
              &\textcolor{black}{37.44$\pm$4.22} &\textcolor{black}{84.39} \\
              
              &&\textcolor{black}{EProxy(S=100)}
              &\textcolor{black}{86.55$\pm$1.52} &\textcolor{black}{96.67}
              &\textcolor{black}{64.16$\pm$2.80} &\textcolor{black}{95.70}
              &\textcolor{black}{34.59$\pm$4.14} &\textcolor{black}{842.76}\\

              &&\textcolor{black}{EProxy(S=1000)}         
              &\textcolor{black}{85.76$\pm$1.05} &\textcolor{black}{988.53}
              &\textcolor{black}{60.74$\pm$5.65} &\textcolor{black}{990.87}
              &\textcolor{black}{34.80$\pm$1.65} &\textcolor{black}{8548} \\
              
              &&\textcolor{black}{Synflow(S=10)} 
              &\textcolor{black}{86.97$\pm$1.46} &\textcolor{black}{0.39} 
              &\textcolor{black}{54.41$\pm$5.40} &\textcolor{black}{0.39}
              &\textcolor{black}{32.08$\pm$4.42} &\textcolor{black}{1.43} \\
              
              &&\textcolor{black}{Synflow(S=100)} 
              &\textcolor{black}{86.13$\pm$1.77} &\textcolor{black}{3.78} 
              &\textcolor{black}{52.07$\pm$4.66} &\textcolor{black}{3.81}  
              &\textcolor{black}{30.74$\pm$3.90} &\textcolor{black}{12.90} \\

              &&\textcolor{black}{Synflow(S=1000)}         
              &\textcolor{black}{86.96$\pm$1.35} &\textcolor{black}{37.25}
              &\textcolor{black}{51.24$\pm$3.83} &\textcolor{black}{37.17}
              &\textcolor{black}{28.85$\pm$3.43} &\textcolor{black}{126.93} \\
              
              &&\textcolor{black}{grad\_norm(S=10)} 
              &\textcolor{black}{88.34$\pm$1.02} &\textcolor{black}{0.38} 
              &\textcolor{black}{66.05$\pm$2.53} &\textcolor{black}{0.38} 
              &\textcolor{black}{39.95$\pm$3.06} &\textcolor{black}{4.26} \\
              
              &&\textcolor{black}{grad\_norm(S=100)}
              &\textcolor{black}{88.52$\pm$1.05} &\textcolor{black}{3.38} 
              &\textcolor{black}{66.75$\pm$1.96} &\textcolor{black}{3.52} 
              &\textcolor{black}{40.23$\pm$2.94} &\textcolor{black}{41.51} \\

              &&\textcolor{black}{grad\_norm(S=1000)}         
              &\textcolor{black}{88.40$\pm$1.04} &\textcolor{black}{34.42}
              &\textcolor{black}{66.85$\pm$1.88} &\textcolor{black}{33.07}
              &\textcolor{black}{40.17$\pm$2.96} &\textcolor{black}{410.43} \\

              &&\textcolor{black}{Snip(S=10)} 
              &\textcolor{black}{89.49$\pm$0.45} &\textcolor{black}{3.00}
              &\textcolor{black}{68.00$\pm$1.33} &\textcolor{black}{3.18}
              &\textcolor{black}{42.81$\pm$2.02} &\textcolor{black}{5.24}\\
              
              &&\textcolor{black}{Snip(S=100)}
              &\textcolor{black}{90.04$\pm$0.26} &\textcolor{black}{60.41}
              &\textcolor{black}{68.97$\pm$0.87} &\textcolor{black}{62.54}
              &\textcolor{black}{44.65$\pm$0.81} &\textcolor{black}{83.22}\\

              &&\textcolor{black}{Snip(S=1000)}         
              &\textcolor{black}{90.31$\pm$0.11} &\textcolor{black}{621.16}
              &\textcolor{black}{69.27$\pm$0.79} &\textcolor{black}{639.71}
              &\textcolor{black}{45.25$\pm$0.32} &\textcolor{black}{827.57}\\
              
\cline{2-9}
              &\multirow{10}{11em}{Layer-based\\Training-free\\(no loss and label required)} &Random                   
              &87.90$\pm$1.29 &N/A
              &62.43$\pm$5.33 &N/A
              &37.13$\pm$4.57 &N/A\\
              
              &&NASWOT(S=10)              
              &88.95$\pm$0.83 &\textcolor{black}{0.31} 
              &64.42$\pm$4.46 &\textcolor{black}{0.31} 
              &40.61$\pm$3.55 &\textcolor{black}{1.08}\\
              
              &&NASWOT(S=100)             
              &89.73$\pm$0.48 &\textcolor{black}{2.94}
              &67.12$\pm$2.91 &\textcolor{black}{2.92}
              &42.94$\pm$2.54 &\textcolor{black}{11.12} \\

              &&\textcolor{black}{NASWOT(S=1000)}         
              &90.10$\pm$0.35 &\textcolor{black}{29.53}
              &69.05$\pm$1.44 &\textcolor{black}{30.62}
              &44.86$\pm$0.84 &\textcolor{black}{76.26} \\
              
              &&\textcolor{black}{TE-NAS(S=10)} 
              &\textcolor{black}{88.26$\pm$0.95} &\textcolor{black}{7.83}
              &\textcolor{black}{61.43$\pm$4.77} &\textcolor{black}{8.19}
              &\textcolor{black}{37.02$\pm$3.75} &\textcolor{black}{156.36}\\
              
              &&\textcolor{black}{TE-NAS(S=100)}
              &\textcolor{black}{88.52$\pm$0.75} &\textcolor{black}{76.75} 
              &\textcolor{black}{54.39$\pm$4.70} &\textcolor{black}{78.62}  
              &\textcolor{black}{36.48$\pm$3.31} &\textcolor{black}{1588} \\

              &&\textcolor{black}{TE-NAS(S=1000)}         
              &\textcolor{black}{85.42$\pm$2.76} &\textcolor{black}{820.74}
              &\textcolor{black}{54.79$\pm$4.36} &\textcolor{black}{794.64}
              &\textcolor{black}{37.39$\pm$1.63} &\textcolor{black}{15946} \\
              
              &&{\bf RBFleX-NAS(S=10)}          
              &\textcolor{black}{{\bf 92.91$\pm$0.33}} &\textcolor{black}{{\bf 0.41}}
              &\textcolor{black}{{\bf 69.10$\pm$0.89}} &\textcolor{black}{{\bf 0.41}} 
              &\textcolor{black}{{\bf 45.66$\pm$7.10}} &\textcolor{black}{{\bf 0.84}}\\
              
              &&{\bf RBFleX-NAS(S=100)}         
              &\textcolor{black}{{\bf 93.02$\pm$0.02}} &\textcolor{black}{{\bf 3.95}} 
              &\textcolor{black}{{\bf 69.21$\pm$1.02}} &\textcolor{black}{{\bf 3.96}} 
              &\textcolor{black}{{\bf 46.06$\pm$0.16}} &\textcolor{black}{{\bf 8.17}}\\

              &&{\bf RBFleX-NAS(S=1000)}         
              &\textcolor{black}{{\bf 93.16$\pm$0.24}} &\textcolor{black}{{\bf 40.51}} 
              &\textcolor{black}{{\bf 70.36$\pm$0.11}} &\textcolor{black}{{\bf 40.15}} 
              &\textcolor{black}{{\bf 46.07$\pm$0.61}} &\textcolor{black}{{\bf 84.03}}\\

\hline

\end{tabular} 
\label{table:benchmark-table}
\end{table*}

\subsection{Analysis of Architecture Search}
\label{Analysis_of_Architecture_Search}
\textcolor{black}{This experiment compares the accuracies and search costs of the best networks (i.e., the networks with the highest score) identified by RBFleX-NAS and other state-of-the-art gradient-based and layer-based training-free NAS methods in NAS-Bench-201, NATS-Bench-SSS, and DARTS design space.}
To evaluate them on NAS-Bench-201 and NATS-Bench-SSS, we randomly sample a total of $S$ networks from each design space.
\textcolor{black}{We perform architecture search following the methodology in NASWOT.} 
\textcolor{black}{Specifically, we firstly work with the CIFAR-10 dataset for NAS-Bench-201 and select the network with the highest score. In the next scoring phase, we evaluate the accuracy of the selected network over CIFAR-10, CIFAR-100, and ImageNet, resulting in identical search time across datasets. 
For NATS-Bench-SSS, networks are selected and scored separately for CIFAR-10, CIFAR-100, and ImageNet. The accuracy of each network is then evaluated on the respective dataset, resulting in different search time across datasets.}
\textcolor{black}{
For  $S=10$  and  $S=100$, the accuracy metrics for ZiCo, Synflow, Grad Norm, and NASWOT are averaged over 500 runs to ensure statistical robustness. However, TE-NAS and EProxy are averaged over 50 runs due to computational limitations.
For  $S=1000$, the accuracy metrics for all gradient-based and layer-based training-free NAS methods are averaged over 50 runs, except for TE-NAS and EProxy, which are averaged over 5 runs, also due to computational limitations. 
}
Besides, we evaluate other common NAS algorithms such as Random Search with Parameter Sharing (RSPS)\cite{NAS_RSPS}, DART\cite{NAS_GA1}, GDAS\cite{NAS_GA5}, ENAS\cite{NAS-Train3}, SETN\cite{NAS_SETN}, TAS\cite{NAS_TAS}, FBNet-v2\cite{NAS_FBNet}, and TuNAS\cite{bender2020can} and compare them with RBFleX-NAS. Specifically, we randomly select 500 networks from their respective design space and calculate the average accuracy.
\textcolor{black}{In the DARTS design space, RBFleX-NAS evaluates all 5,000 networks. We train the top-1 scored network for 800 epochs with a batch size of 96 to align with the evaluation method used by the benchmark algorithms.} 
The remaining parameters are configured the same as those in the DARTS papers\cite{NAS_GA1}. Besides, we also evaluate ZiCo\cite{li2023zico}, TE-NAS\cite{chen2021neural}, P-DARTS\cite{chen2019progressive}, PC-DARTS\cite{NAS_GA7}, ENAS\cite{NAS-Train3}, GDAS\cite{NAS_GA5}, DARTS-v1\cite{NAS_GA1}, and PNAS\cite{liu2018progressive}.
\textcolor{black}{We employ a single NVIDIA Tesla V100 GPU for all computations to implement this experiment.}

\begin{figure}
\centering
\includegraphics[width=0.48\textwidth]{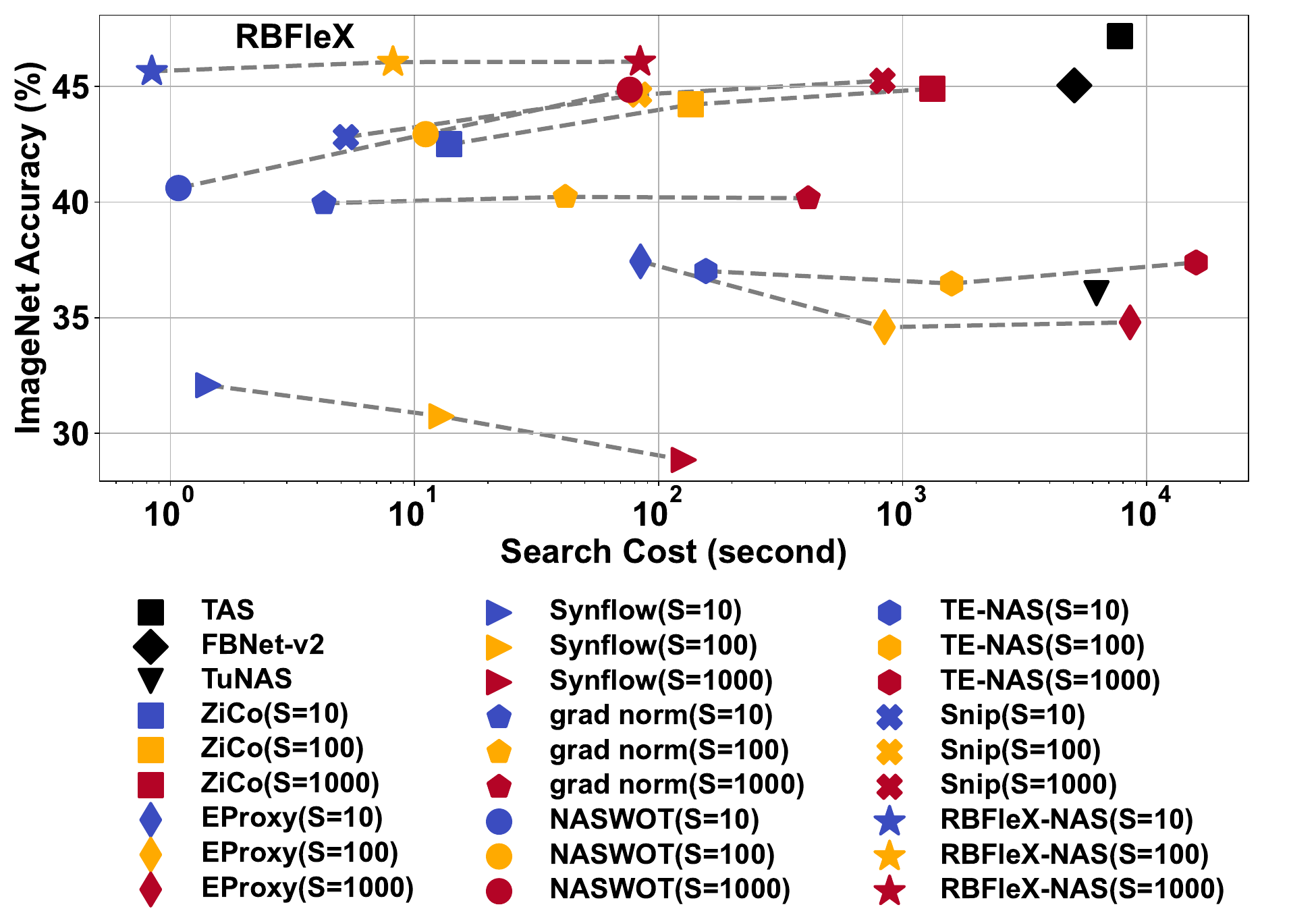} 
\caption{\textcolor{black}{Search cost versus network accuracy in ImageNet classification on NATS-Bench-SSS.}
}
\label{fig:Time-Precision}
\end{figure}

Table \ref{table:benchmark-table} lists the results in detail. \textcolor{black}{Specifically, in NAS-Bench-201 with a sample size of 1000, RBFleX-NAS exhibits superior performance compared to all other state-of-the-art NAS methods, achieving the highest top-1 accuracy of 93.30\% over CIFAR-10, 70.99\% over CIFAR-100, and 44.57\% over ImageNet, respectively.}
Moreover, our results outperform NASWOT in terms of a smaller variance as Table \ref{table:benchmark-table} shows (i.e., $0.35/0.21/0.70$ in RBFlex-NAS vs. $0.84/1.45/2.52$ in NASWOT).
\textcolor{black}{In NATS-Bench-SSS with a sample size of 1000, our searched top-1 accuracy is 93.16\% over CIFAR-10, 70.36\% over CIFAR-100, and 46.07\% over ImageNet, respectively, higher than the accuracy resulting from all other training-free NAS algorithms.}
\textcolor{black}{Fig.\ref{fig:Time-Precision} further compares the search cost with respect to network accuracy in NATS-Bench-SSS, showing that RBFleX-NAS can identify networks with higher accuracy yet with less search time in every sample size.}

\begin{figure}
\centering
\includegraphics[width=0.45\textwidth]{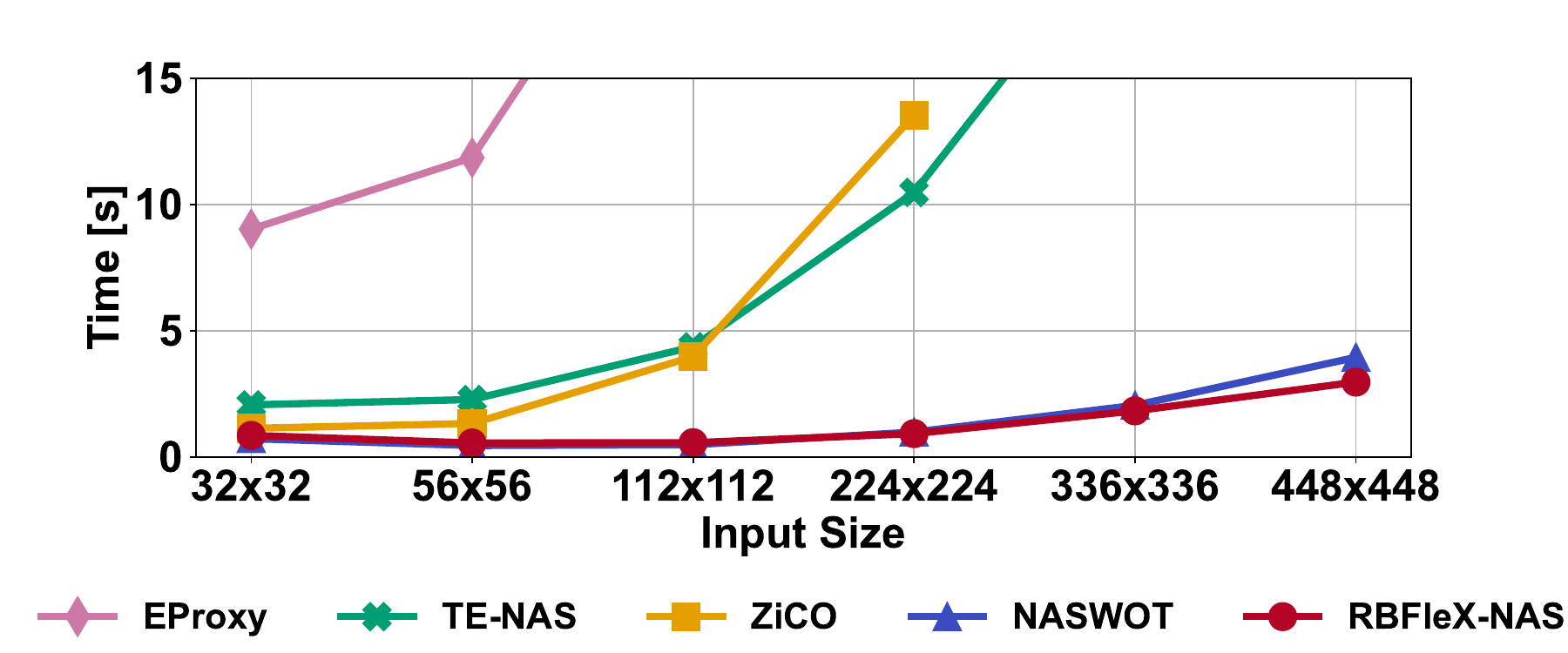} 
\caption{\textcolor{black}{Comparison of search cost (S=10) vs. input size for different NAS methods. ZiCO is unable to run on inputs larger than 336$\times$336 due to insufficient GPU memory (NVIDIA Tesla V100 32GB).}}
\label{fig:Time-Inputs}
\end{figure}

\textcolor{black}{Fig.\ref{fig:Time-Inputs} presents the averaged search cost for $S=10$ over five runs for various state-of-the-art training-free NAS methods (i.e., NASWOT, EProxy, TE-NAS, and ZiCo) across different input sizes. Notably, RBFleX-NAS maintains a relatively low and stable search cost as input size increases, outperforming other methods in efficiency.
At the largest input size (448$\times$448), RBFleX-NAS achieves a search cost of only 2.97 seconds, significantly faster than TE-NAS (39.34 seconds), EProxy (346.2 seconds), and NASWOT (3.95 seconds). Additionally, ZiCo fails to operate with an input size larger than 336$\times$336  due to the GPU memory limitation.}

\begin{figure}
\centering
\includegraphics[width=0.45\textwidth]{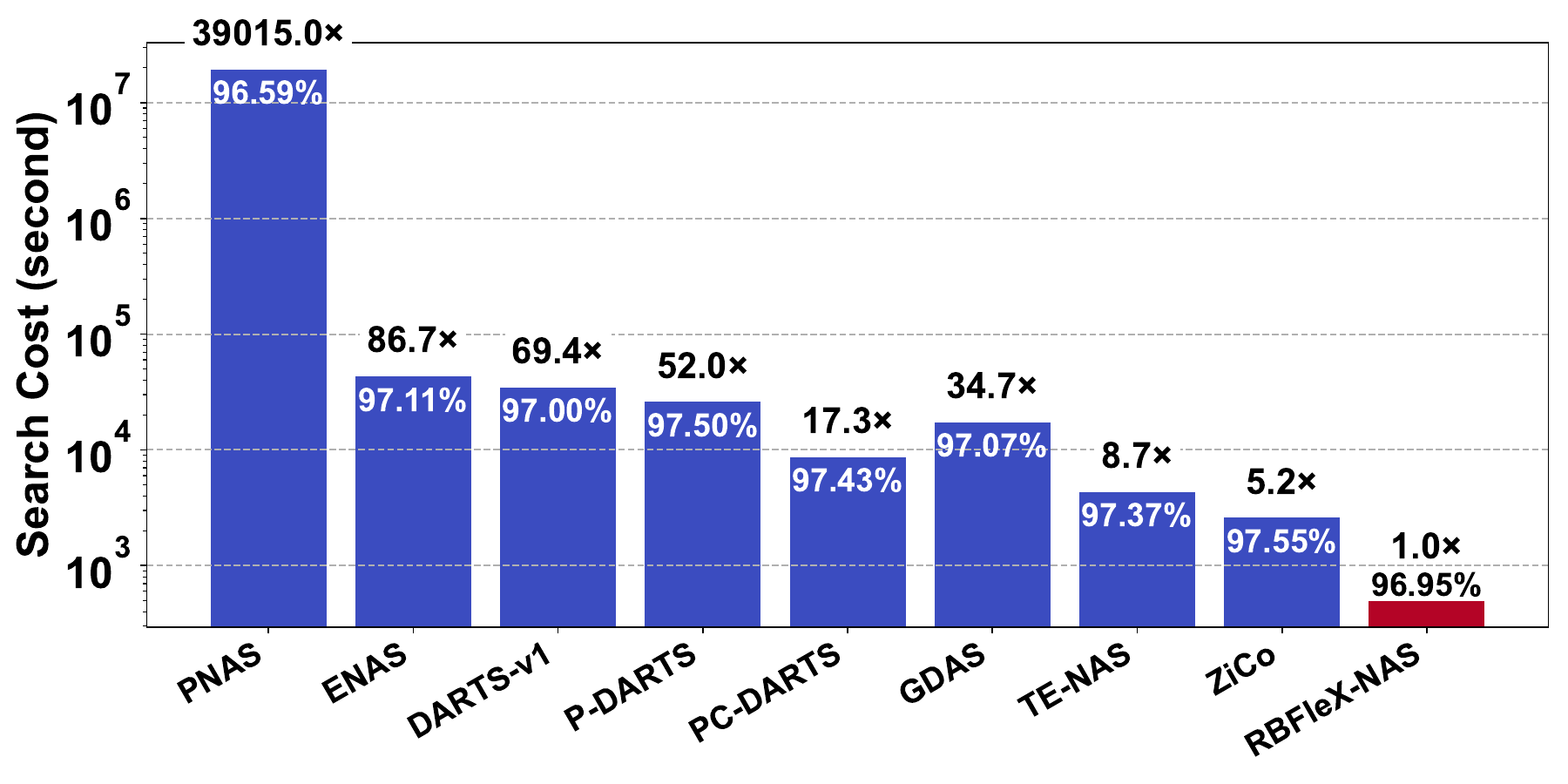} 
\caption{\textcolor{black}{Comparison of search cost vs. top-1 CIFAR-10 accuracy on DARTS design space. The red dot indicates RBFleX-NAS, running 5.2$\times$ faster than ZiCo to search the 96.95\%-accuracy network.}
}
\label{fig:time-Darts}
\end{figure}

\textcolor{black}{Fig.\ref{fig:time-Darts} 
compares different NAS methods in the DARTS design space using network accuracy over CIFAR-10 and search cost as metrics. While ZiCo achieves the highest accuracy of 97.55\%, RBFleX-NAS closely follows with an accuracy of 96.95\%. Notably, ZiCo runs 5.2$\times$ slower than RBFlex-NAS with only a marginal accuracy improvement of 0.6\%.
}

\subsection{Neural Network Activation Function Benchmark (NAFBee) and Evaluation of Non-ReLU Activations}
\textcolor{black}{In this experiment, we perform a comparative analysis among our RBFleX-NAS, NASWOT\cite{NASWOT}, ZiCo\cite{li2023zico}, TE-NAS\cite{chen2021neural}, grad\_norm\cite{abdelfattah2021zerocost}, snip\cite{lee2018snip}, and synflow\cite{NEURIPS2020_46a4378f} on a design space with activation functions.} To accomplish this, we have created the Neural Network Activation Function Benchmark (NAFBee). NAFBee utilizes VGG-19 and BERT\textsubscript{BASE}\cite{devlin2019bert} network as backbones. We replace all activation functions on a backbone with various types of activation functions and build architecture candidates. We select widely used activation functions such as ReLU, GELU\cite{GELU}, SiLU\cite{SiLU}, LeakyReLU, ReLU6, Mish\cite{Mish}, Hardswish, CELU\cite{CELU}, ELU\cite{ELU}, Hardtanh, and SELU\cite{SELU}. Therefore, NAFBee (VGG-19) and NAFBee (BERT) each have 11 architecture candidates.

We then acquire the ground-truth accuracy of each candidate network in NAFBee (VGG-19) and NAFBee (BERT) via training. For NAFBee (VGG-19), the 200-epoch training deploys the SGD optimizer with a cosine decay learning rate from 0.1 to 0, a momentum of 0.9, a weight decay of 5e-4 and a batch size of 128. This configuration is inspired by NATS-Bench\cite{NATS-Bench}. The target dataset to work with NAFBee (VGG-19) is CIFAR-10 which has been preprocessed by randomly cropping to 32×32 pixels with 4-pixel padding on each border as well as horizontal random flipping and normalizing across the RGB channels. For NAFBee (BERT), we utilize the same fine-tuning parameters as in the BERT paper \cite{devlin2019bert}. The target dataset for demonstration is Stanford Sentiment Treebank (SST-2)\cite{sst-2}. SST-2 incorporates a total of 215,154 unique phrases for binary classification.
RBFleX-NAS and the training-free NAS algorithms under comparison evaluate each network configuration with the aforementioned activation functions.
In the evaluation phase, we compute the correlation coefficients between the ground-truth accuracy and the corresponding scores over CIFAR-10 using VGG-19 and over SST-2 using BERT. The batch size $N$ for RBFleX-NAS and NASWOT is 16. The number of networks $M$ for HDA is 1.

\begin{table*}
\centering
\footnotesize
\caption{The Score Result in NAFBee(VGG-19). Pearson and Kendall Correlation Between Validated Accuracy and Score.}
\begin{tabular}{c|lc|ccccccc}
\hline
\multirow{3}{6em}{Design Space} &\multirow{3}{3em}{Activation Functions}&\multirow{3}{6em}{Accuracy(\%)} &\multicolumn{7}{c}{Score}\\
\cline{4-10}
& & &\multicolumn{3}{c|}{Layer-based}  &\multicolumn{4}{c}{Gradient-based} \\
& & &{\bf RBFleX-NAS} &NASWOT &\multicolumn{1}{l|}{\textcolor{black}{TE-NAS}} &grad\_norm &snip &synflow &\textcolor{black}{ZiCo}\\
\hline\hline

\multirow{11}{6em}{NAFBee\\(VGG-19)} 

&ReLU	
&{\bf 91.06}
&\textcolor{black}{\bf -6190.45}	&191.72 &\textcolor{black}{-2.11}
&573.13	&3569.66 &5.19E+22 &\textcolor{black}{259.02}
\\

&GELU\cite{GELU}
&91.03	
&\textcolor{black}{-6207.12}	&{\bf 192.04} &\textcolor{black}{-4.45}
&478.64	&3045.44	&5.25E+22 &\textcolor{black}{257.28}
\\

&SiLU\cite{SiLU}
&91.03	
&\textcolor{black}{-6212.32}	&191.88 &\textcolor{black}{-9.33}
&329.59	&2027.23	&5.01E+22 &\textcolor{black}{257.98}
\\

&LeakyReLU
&90.75	
&\textcolor{black}{-6195.74}	&191.75 &\textcolor{black}{-2.10}
&560.38	&3547.21	&5.19E+22 &\textcolor{black}{257.89}
\\

&ReLU6
&90.61	
&\textcolor{black}{-6194.21}	&191.62 &\textcolor{black}{-2.14}
&{\bf 574.87}	&{\bf 3574.24}	&6.79E+02 &\textcolor{black}{\bf 259.98}
\\

&Mish\cite{Mish}
&90.54	
&\textcolor{black}{-6202.14}	&192.00 &\textcolor{black}{-2.43}
&426.65	 &2810.60	&5.19E+22 &\textcolor{black}{258.78}
\\

&Hardswish
&89.94	
&\textcolor{black}{-6224.58}	&191.80 &\textcolor{black}{-181.95}
&473.98	&2830.19	&4.85E+22 &\textcolor{black}{257.09}
\\

&CELU\cite{CELU}
&88.36	
&\textcolor{black}{-6216.53}	&191.95 &\textcolor{black}{-1.62}
&302.95	&2126.31	&5.42E+22 &\textcolor{black}{257.35}
\\

&ELU\cite{ELU}
&87.01	
&\textcolor{black}{-6215.80}	&191.84 &\textcolor{black}{{\bf -1.61}}
&313.30	&2182.29	&5.17E+22 &\textcolor{black}{259.02}
\\

&Hardtanh
&86.74	
&\textcolor{black}{-6216.71}	&192.00 &\textcolor{black}{-2.08}
&566.95	&3501.04	&114.17 &\textcolor{black}{258.33}
\\

&SELU\cite{SELU}
&85.83	
&\textcolor{black}{-6217.89}	&191.97 &\textcolor{black}{-1.64}
&345.97	&2404.90	&{\bf 1.12E+23} &\textcolor{black}{259.57}
\\

\hline\hline

\multicolumn{3}{c|}{Pearson Correlation } 
&\textcolor{black}{0.623} &-0.384 &\textcolor{black}{-0.117}
&0.708 &-0.417 &-0.574 &\textcolor{black}{-0.255}\\
\multicolumn{3}{c|}{Kendall Correlation }
&\textcolor{black}{0.550} &-0.240 &\textcolor{black}{-0.367}
&0.257 &0.220 &0.019 &\textcolor{black}{-0.110}\\

\hline

\end{tabular}
\label{table:Activation Layer VGG}
\end{table*}

Table \ref{table:Activation Layer VGG} shows the experimental results obtained using NAFBee (VGG-19). The ReLU activation function achieves the highest performance with an accuracy of 91.06\%. RBFleX-NAS successfully identifies this optimal network configuration with the highest score, whereas other training-free NAS fail to achieve it. 

\begin{table*}
\centering
\footnotesize
\caption{The Score Result in NAFBee(BERT). Pearson and Kendall Correlation Between Validated Accuracy and Score. 
}
\begin{tabular}{c|lc|ccccccc}
\hline
\multirow{3}{6em}{Design Space}&\multirow{3}{3em}{Activation Functions}&\multirow{3}{6em}{Accuracy(\%)} &\multicolumn{7}{c}{Score}\\
\cline{4-10}
& & &\multicolumn{3}{c|}{Layer-based} &\multicolumn{4}{c}{Gradient-based} \\
& & &{\bf RBFleX-NAS} &NASWOT &\multicolumn{1}{l|}{\textcolor{black}{TE-NAS}} 
&grad\_norm &snip &synflow &\textcolor{black}{ZiCo}\\
\hline\hline

\multirow{11}{6em}{NAFBee\\(BERT)}
&GELU &{\bf 92.97}	
&{\bf -4388.62}	&198.05	&\textcolor{black}{\bf -16.09}
&194.23	&2441.72	&5.07e-13 &\textcolor{black}{1193}\\

&ReLU	&92.26	
&-5635.96	&174.57	&\textcolor{black}{-159.27}
&36.45	&193.71	&2.68e-13 &\textcolor{black}{1161}\\

&LeakyReLU	&91.76	
&-5406.85	&181.83	&\textcolor{black}{-133.37}
&50.05	&589.93	&2.36e-13 &\textcolor{black}{\bf 1231}\\

&ReLU6	&91.60	
&-5664.52	&173.18	&\textcolor{black}{-173.63}
&17.18	&155.81	&1.09e-12 &\textcolor{black}{1188}\\

&SiLU	&86.77	
&-4516.79	&200.1	&\textcolor{black}{-44.06}
&238.28	&{\bf 8276.62}	&8.77e-13 &\textcolor{black}{1198}\\

&Hardswish	&85.23	
&-4460.36	&{\bf 203.95}	&\textcolor{black}{-34.41}
&418.03	&4980.75	&8.51e-13 &\textcolor{black}{1200}\\

&Mish	&84.51	
&-4437.43	&201.51	&\textcolor{black}{-28.62}
&{\bf 547.23}	&4268.57	&8.81e-13 &\textcolor{black}{1183}\\

&ELU	&49.92	
&-7638.97	&176.97	&\textcolor{black}{-110.74}
&2.73	&144.42	&8.06e-13 &\textcolor{black}{1193}\\

&Hardtanh	&49.92	
&-7621.49	&176.72	&\textcolor{black}{-120.74}
&22.66	&10.89	&4.37e-13 &\textcolor{black}{1168}\\

&SELU	&49.92	
&-7482.86	&173.16	&\textcolor{black}{-147.15}
&11.81	&24.95	&1.02e-12 &\textcolor{black}{1169}\\

&CELU	&49.92	
&-7637.44	&177.17	&\textcolor{black}{-134.80}
&10.05	&58.23	&{\bf 1.82e-12} &\textcolor{black}{1176}\\

\hline\hline

\multicolumn{3}{c|}{Pearson Correlation } 
&0.906 &0.484 &\textcolor{black}{0.306}
&0.433 &0.436 &-0.434 &\textcolor{black}{0.425}\\
\multicolumn{3}{c|}{Kendall Correlation }
&0.443 &0.135 &\textcolor{black}{0.058}
&0.289 &0.443 &-0.289 &\textcolor{black}{0.250}\\

\hline

\end{tabular}
\label{table:Activation Layer BERT}
\end{table*}

Table \ref{table:Activation Layer BERT} shows the scores and correlation values by different training-free proxies using NAFBee (BERT). The BERT model with GELU activation function achieves the highest accuracy of 92.97\%. RBFleX-NAS outperforms both layer-based and gradient-based NAS as it successfully identifies the best-performing network during activation function search. \textcolor{black}{In addition, the Kendall correlation of RBFleX-NAS when scoring the network with GELU is 0.443 while the correlation of TE-NAS is merely 0.058. Apparently, RBFleX-NAS can accurately evaluate networks incorporating various widely adopted activation functions, expanding the design space with activation function exploration for network architecture search.}

\section{Conclusion}
\label{sec:conclusion}
This work introduces RBFleX-NAS, a training-free NAS approach that leverages a Radial Basis Function (RBF) kernel with a hyperparameter detection algorithm. RBFleX-NAS evaluates candidate networks based on the similarity of activation outputs and input feature maps from the last layer of a network across different input images. It demonstrates superior top-1 accuracy in network search and excellent Kendall correlation across diverse design spaces in image classification, object classification/segmentation, and natural language processing tasks.

Additionally, we present NAFBee, a novel benchmarking design space that supports a broader range of activation functions, enabling activation function exploration. Compared to state-of-the-art training-free NAS methods, RBFleX-NAS accurately identifies the best-performing networks with activation function searches for architectures such as VGG-19 and BERT. Furthermore, it offers rapid network evaluation, requiring only seconds to assess a candidate network. This makes it a highly efficient tool for identifying optimal architectures for diverse AI applications.


\bibliographystyle{ieeetr}

\vspace{11pt}

\begin{IEEEbiography}[{\includegraphics[width=1in,height=1.25in,clip,keepaspectratio]{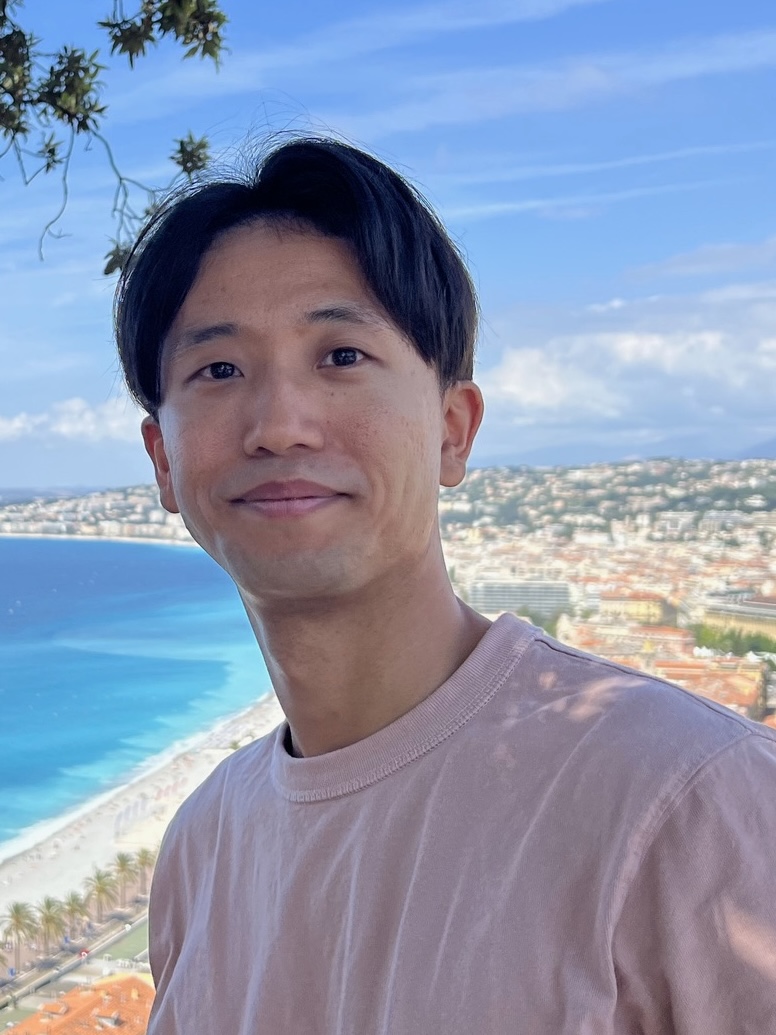}}]{Tomomasa Yamasaki}
received his B.Eng. and M.Eng. degrees in System Engineering from Aoyama Gakuin University, Tokyo, Japan. He is currently pursuing a Ph.D. in Computer Science at the Singapore University of Technology and Design (SUTD), Singapore. His research interests focus on neural architecture search (NAS), and design automation tools for hardware–software co-optimization.

He received the Best Paper Award at the ACM/IEEE International Symposium on Low Power Electronics and Design (ISLPED) in 2023. He was also a recipient of the Best Student Paper Award at the 8th IIAE International Conference on Industrial Application Engineering in 2020.

\end{IEEEbiography}

\begin{IEEEbiography}[{\includegraphics[width=1in,height=1.25in,clip,keepaspectratio]{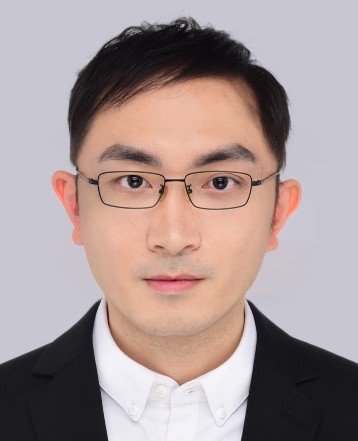}}]{Zhehui Wang}
received B.S. degree in Electrical Engineering from Fudan University, China, in 2010, and Ph.D. degree in Electronic and Computer Engineering from Hong Kong University of Science and Technology, Hong Kong, in 2016. He is currently a Research Scientist with the Institute of High Performance Computing, Agency for Science, Technology and Research,
Singapore. He authored and co-authored more than 60 research papers in peer-reviewed journals, conferences, and books. His research interests include efficient AI deployment, AI on emerging technologies, hardware-software co-design, and high-performance computing.
\end{IEEEbiography}

\begin{IEEEbiography}[{\includegraphics[width=1in,height=1.25in,clip,keepaspectratio]{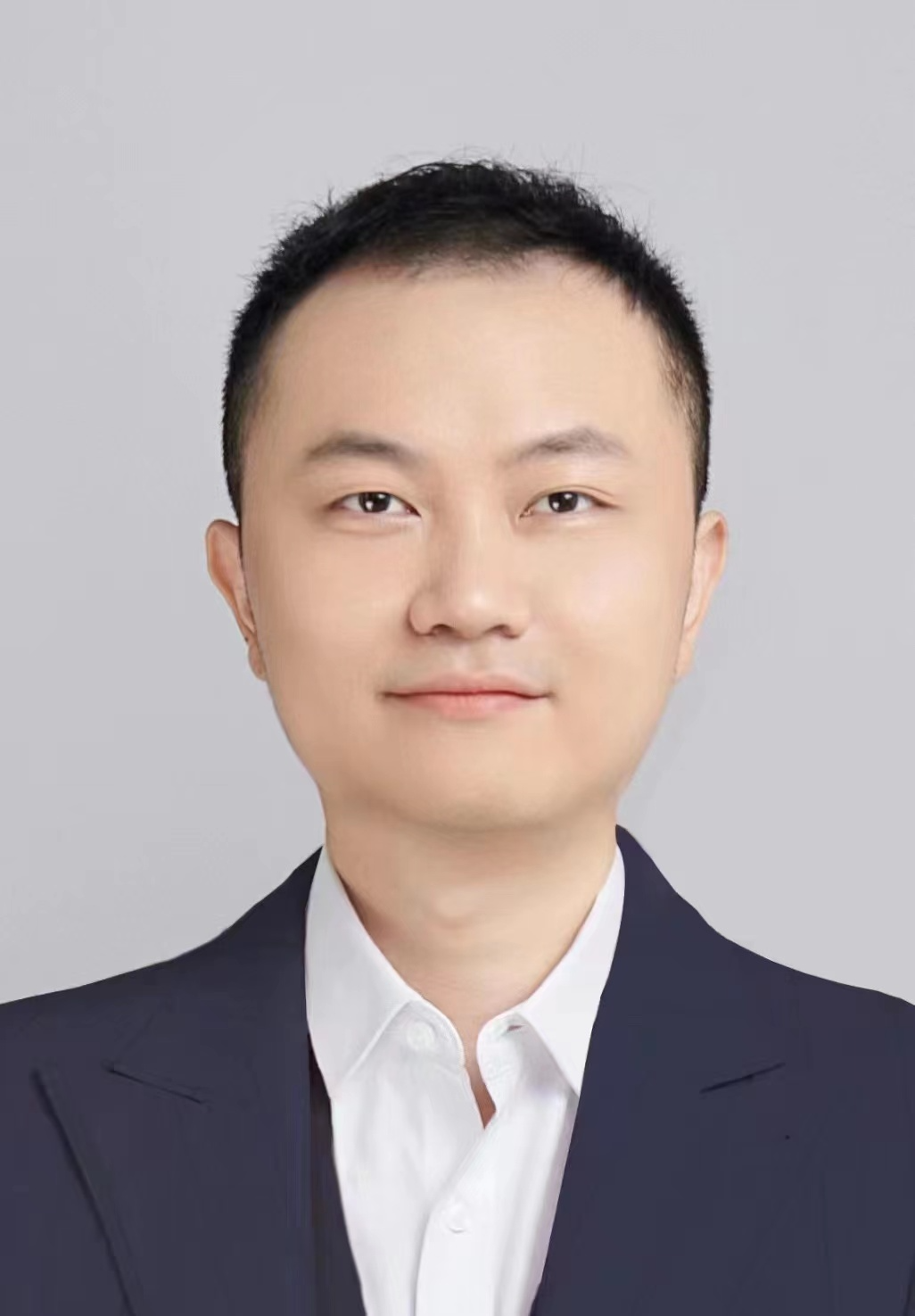}}]{Luo Tao}
received his bachelor’s degree from the Harbin Institute of Technology, Harbin, China, in 2010, his master’s degree from the University of Electronic Science and Technology of China, Chengdu, China, in 2013, and his Ph.D. degree from the School of Computer Science and Engineering, Nanyang Technological University, Singapore, in 2018. He is currently a senior research scientist with the Institute of High Performance Computing (IHPC), Agency for Science, Technology and Research, Singapore (A*STAR), Singapore. His current research interests include high-performance computing,  machine learning, computer architecture, hardware–software co-exploration, quantum computing, efficient AI and its application.
\end{IEEEbiography}

\begin{IEEEbiography}[{\includegraphics[width=1in,height=1.25in,clip,keepaspectratio]{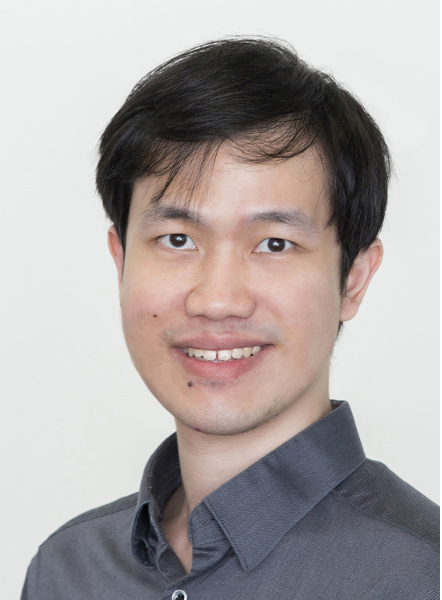}}]{Niangjun Chen}
received the B.S. degree in Computer Science from the University of Cambridge and the M.S. and Ph.D. degrees in Computer Science from the California Institute of Technology. After graduating in 2017, he joined the Institute of High Performance Computing as a Research Scientist working on optimizing logistics and modeling and simulation of the transport systems. Since September 2020, he has been an Assistant Professor with the Singapore University of Technology and Design. He has a joint appointment with the Institute for High Performance Computing, Agency for Science, Technology, and Research. His research interests include optimization, machine learning, game theory, and their applications to complex systems such as smart grids, data centers, and transport.
\end{IEEEbiography}

\begin{IEEEbiography}[{\includegraphics[width=1in,height=1.25in,clip,keepaspectratio]{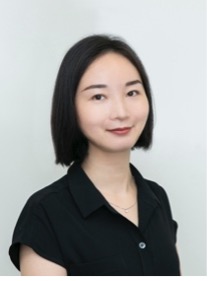}}]{Bo Wang} received the Ph.D. degree in Electrical and Electronic Engineering from Nanyang Technological University, Singapore, in 2015. From 2015 to 2016, she worked as a Staff Circuit Design Engineer at MediaTek, Singapore. She then joined the National University of Singapore as a Post-Doctoral Research Fellow from 2016 to 2020. Since 2020, she has been an Assistant Professor at the Singapore University of Technology and Design, Singapore. She has authored and co-authored many papers published in prestigious journals and conference proceedings, including JSSC, TCAS–I, TVLSI, TCAS–II, A-SSCC, DAC, DATE, ISLPED, and MobiSys. Her research interests span various aspects of energy-efficient system, architecture and circuit design, as well as design automation tools for hardware–software co-optimization.

Dr. Wang received the Distinguished Design Award at the IEEE Asian Solid-State Circuits Conference (A-SSCC) in 2023. She was also a recipient of the Best Paper Award at the ACM/IEEE International Symposium on Low Power Electronics and Design (ISLPED) in 2023 and the International SoC Design Conference (ISOCC) in 2024 and 2014. She currently serves as an Associate Editor for the IEEE Open Journal of Circuits and Systems and has served as a Guest Editor for Frontiers in Neuroscience. 

\end{IEEEbiography}

\vfill

\end{document}